\pdfoutput=1

\documentclass[11pt]{article}

\usepackage{ACL2023}

\usepackage{times}
\usepackage{latexsym}
\usepackage{hyperref}

\usepackage[T1]{fontenc}

\usepackage[utf8]{inputenc}

\usepackage{microtype}

\usepackage{inconsolata}

\usepackage{amsmath}
\usepackage{amssymb}

\usepackage{subcaption}
\usepackage{graphicx}

\usepackage{multirow}

\usepackage{enumitem}

\definecolor{Green}{rgb}{0.0, 0.5, 0.0}
\definecolor{red-brown}{rgb}{0.8, 0.1, 0.1}

\DeclareMathOperator*{\layernorm}{LayerNorm}
\DeclareMathOperator*{\attn}{Attn}
\DeclareMathOperator*{\ffn}{FFN}
\DeclareMathOperator*{\softmax}{Softmax}

\DeclareMathOperator*{\pooling}{Pooling}

\newcommand{\ante}[1]{\textbf{\textcolor{blue}{#1}}}
\newcommand{\pronoun}[1]{\textbf{\textcolor{blue}{#1}}}
\newcommand{\correct}[1]{\textbf{\textcolor{Green}{#1}}}
\newcommand{\incorect}[1]{\textbf{\textcolor{red-brown}{#1}}}

\title{Sequence Shortening for Context-Aware Machine Translation}

\author{Paweł Mąka \and Yusuf Can Semerci \and Jan Scholtes \and Gerasimos Spanakis\\
        Department of Advanced Computing Sciences \\ 
        Maastricht University \\ 
        \texttt{\{pawel.maka, y.semerci, j.scholtes, jerry.spanakis\}@maastrichtuniversity.nl}}

\begin{document}
\maketitle
\begin{abstract}

Context-aware Machine Translation aims to improve translations of sentences by incorporating surrounding sentences as context.  Towards this task, two main architectures have been applied, namely single-encoder (based on concatenation) and multi-encoder models. In this study, we show that a special case of multi-encoder architecture, where the latent representation of the source sentence is cached and reused as the context in the next step, achieves higher accuracy on the contrastive datasets (where the models have to rank the correct translation among the provided sentences) and comparable BLEU and COMET scores as the single- and multi-encoder approaches. Furthermore, we investigate the application of Sequence Shortening to the cached representations. We test three pooling-based shortening techniques and introduce two novel methods - Latent Grouping and Latent Selecting, where the network learns to group tokens or selects the tokens to be cached as context. Our experiments show that the two methods achieve competitive BLEU and COMET scores and accuracies on the contrastive datasets to the other tested methods while potentially allowing for higher interpretability and reducing the growth of memory requirements with increased context size.

\end{abstract}

\section{Introduction}

Following the introduction of the Transformer model \citep{vaswani2017attention}, Sentence-level Machine Translation, where the task is to translate separate sentences, has seen great success in recent years \citep{vaswani2017attention, hassan2018achieving, costa2022no, tiedemann2022democratizing}. However, real-world applications of the translation systems are often used to translate a whole document or a longer discourse (e.g. a transcribed speech). In those circumstances, Sentence-level Machine Translation processes each sentence separately and is incapable of leveraging the surrounding or previous sentences (referred to as the context sentences). This is in contrast to the Context-aware Machine Translation where the context sentences are available to the system. The information in the previous sentences can be helpful to maintain the coherence of the translation and to resolve ambiguities \citep{agrawal2018contextual, bawden-etal-2018-evaluating, muller-etal-2018-large, voita-etal-2019-good}. Both the sentences of the text in the source language and the previously translated sentences can be used as context. The former is referred to as source-side context and the latter as target-side context.

Many Context-aware Machine Translation approaches have been proposed including novel architectures that can be broadly categorized into \textit{single-encoder} and \textit{multi-encoder} types. In single-encoder architectures, the context sentences are concatenated with the current sentence and processed as a long sequence by a single encoder \citep{tiedemann-scherrer-2017-neural, agrawal2018contextual, ma-etal-2020-simple, zhang-etal-2020-long, majumde2022baseline}. In multi-encoder architectures, the context sentences are processed by a separate encoder than the current sentence \citep{tu-etal-2017-context, bawden-etal-2018-evaluating, miculicich-etal-2018-document, maruf-etal-2019-selective, huo-etal-2020-diving, zheng2021towards}. Several multi-encoder approaches \citep{voita-etal-2018-context, li-etal-2020-multi-encoder} involve sharing parameters of encoders. This approach reduces the number of parameters and could also increase the speed of translation when translating the whole document sentence-by-sentence. Inspired by this idea, we investigate multi-encoder architectures where all the encoder parameters are shared \citep{tu-etal-2018-learning, voita-etal-2019-good, wu2022study}, which allows caching the hidden representation of the current sentence and reusing it as the hidden representation of the context when translating subsequent sentences. In this study, we refer to this architecture as \textit{caching}. We experimentally show that this architecture can achieve comparable results to single- and multi-encoder architectures and is more stable in the realm of larger context sizes.

In Transformers, the number of tokens does not change during the processing of the sequence through the encoder (and decoder) layers. Concurrent to Machine Translation, several techniques have been proposed to shorten the sequence of tokens in the task of language modeling \citep{subramanian2020multi, dai2020funnel, nawrot-etal-2022-hierarchical}. In particular, the tokens are combined in the shortening modules that are added between a specified number of encoder layers. Sequence Shortening can lead to the reduction of the computational and memory requirements in the subsequent layers as the requirements of the self-attention module grow quadratically with the number of tokens (although a substantial amount of research is done to mitigate that \citep{kitaev2020reformer, wang2020linformer}).

In this paper, we investigate the application of Sequence Shortening to Context-aware Machine Translation. Specifically, we apply the shortening of the cached hidden representations of the context sentences in the caching multi-encoder architectures. The intuition behind this approach is that a compressed representation of the previously seen sentences should be enough to use as a context while possibly decreasing the computational and memory requirements during inference. Sequence Shortening can be seen as related to the concept of \textit{chunking} from psychology \citep{miller1956magical, terrace2002comparative, mathy2012whats}. To limit the scope, we consider only the source-side context. Additionally, we introduce \textit{Latent Grouping} and \textit{Latent Selecting} - new shortening techniques where the network can learn how to group or select tokens to form a shortened sequence. Our experiments indicate that sequence shortening can be leveraged to improve the stability of training for larger context sizes (we tested up to 10 previous sentences as context) while achieving comparable results for smaller context sizes.

\section{Related Work}

\subsection{Context-aware Machine Translation}
 
A straightforward approach to incorporate context into Machine Translation is to concatenate previous sentences with the current sentence, which has been referred to as \textit{concatenation} or \textit{single-encoder} architecture because only a single encoder is used \citep{tiedemann-scherrer-2017-neural, ma-etal-2020-simple, zhang-etal-2020-long}. This architecture has achieved very good results \citep{majumde2022baseline} even on long context sizes (of up to 2000 tokens) when data augmentation was used \citep{sun-etal-2022-rethinking} but even longer context sizes will result in a sharply increasing memory and computational complexity \citep{feng-etal-2022-learn}. The \textit{multi-encoder} approach is to encode the context sentences by a separate encoder \citep{jean2017does, miculicich-etal-2018-document, maruf-etal-2019-selective, huo-etal-2020-diving, zheng2021towards}. While the encoders are separate in multi-encoder architectures, weight-sharing between them has been investigated in previous works \citep{voita-etal-2018-context, tu-etal-2018-learning, li-etal-2020-multi-encoder, wu2022study}. Existing studies also investigated hierarchical attention \citep{miculicich-etal-2018-document, bawden-etal-2018-evaluating, wu2022study, chen2022one}, sparse attention \citep{maruf-etal-2019-selective, bao-etal-2021-g}, aggregating the hidden representation of the context tokens \citep{morishita-etal-2021-context}, and post-processing the translation \citep{voita-etal-2019-good, voita-etal-2019-context}. Similar to ours, several works use a memory mechanism \citep{feng-etal-2022-learn, bulatov2022recurrent}. The main differences are that the memory-based techniques rely on the attention mechanism to collect information from the sentences. In addition to that, our method allows the tokens in the current sentence to work as a hub tokens instead of the learned (but fixed) tokens of the memory in the initial step or the memory vectors from the previous steps. In the memory approaches, the number of tokens is constant while in the models employing shortening the number of tokens is dependent on the number of context segments.

Mostly orthogonal to architectural approaches, another line of work concentrates on making the models use the context more effectively. These methods utilize regularization such as dropout of the tokens in the source sentence \citep[CoWord dropout;][]{fernandes-etal-2021-measuring}, attention regularization based on human translators \citep{yin-etal-2021-context}, and data augmentation \citep{lupo-etal-2022-divide} along with contrastive learning \citep{hwang-etal-2021-contrastive}.

It has been argued that widely used sentence-level metrics (such as BLEU \citep{papineni-etal-2002-bleu}) are ill-equipped to measure the translation quality with regard to the inter-sentential phenomena \citep{hardmeier2012discourse, wong-kit-2012-extending}. For this reason, research has been done to measure the usage of context by machine translation models, where two main avenues have been explored: introducing new metrics \citep{fernandes-etal-2021-measuring, fernandes-etal-2023-translation} and contrastive datasets \cite{muller-etal-2018-large, bawden-etal-2018-evaluating, voita-etal-2019-good, lopes-etal-2020-document}. In the contrastive datasets, the model is presented with the task of ranking several translations of the same source sentence with the same context. The provided translations differ only partially and the provided context is required to choose the correct translation.

\subsection{Sequence Shortening}

Sequence Shortening has been introduced as a way to exploit the hierarchical structure of language to reduce the memory and computational cost of the Transformer architecture \citep{subramanian2020multi, dai2020funnel, nawrot-etal-2022-hierarchical}. Shortening can be done by average pooling of the hidden representation of the tokens \cite{subramanian2020multi}. Allowing the tokens of the shortened sequence to attend to the hidden representation of the original sequence was found beneficial \citep{dai2020funnel}. Replacing average pooling with the linear transformation of the concatenated representation of the tokens of the original sequence has also been used \citep{nawrot-etal-2022-hierarchical}. Another way of shortening the sequence is to find and retain only the most important tokens of the original sequence \cite{goyal2020power}. Furthermore, a large body of work improve the context size or the efficiency of the Transformer model \citep{beltagy2020longformer, kitaev2020reformer, dai-etal-2019-transformer} which has been referenced in comprehensive surveys \citep{tay2022efficient, lin2022survey}.

The work that is architecturally most closely related to one of our methods \textit{Latent Grouping} is the Charformer \citep{tay2021charformer} architecture, where the tokenization is performed by a sub-network that learns to select block sizes for characters of the input sequence. The block size representations are subsequently summed with weights predicted by the sub-network. \textit{Latent Grouping} differs from Charformer in the placement of the grouping (after the encoder in the case of \textit{Latent Grouping}) and the aggregated representation (encoder representations of tokens themselves in the case of \textit{Latent Grouping}).

Our work lies in the intersection of Context-aware Machine Translation and Sequence Shortening. We test the performance of caching architecture against single- and multi-encoder architectures and investigate applying shortening to the cached sentences. 

\section{Background}

\subsection{Transformer}

The Transformer architecture, introduced for sentence-level translation, consists of the encoder and decoder \citep{vaswani2017attention}. The sentence in the source language is tokenized and embedded before it is passed to the encoder. The encoder processes the sequence by $L$ consecutive encoder layers, each consisting of the self-attention module and the element-wise feed-forward network. Residual connection is added around both modules followed by Layer Normalization \citep{ba2016layer}. 

Hidden representation of the $L$-th encoder layer $H^L$ is fed into the decoder, which auto-regressively produces the output sequence $Y=(y_1, ..., y_T),$ until it reaches the end-of-sequence token. Decoder layers process the currently produced sequence with the self-attention module, followed by the cross-attention module and feed-forward network. Unlike in the encoder, the self-attention module in the decoder uses causal masking (the tokens can not attend to the future tokens). In Cross-attention, multi-head attention uses the decoded sequence as queries and the encoder output as keys and values. Residual connection and Layer Normalization are applied after each module.

\subsection{Pooling-based Shortening}

Sequence Shortening is a method that results in a reduction in the number of tokens in a sequence by combining the tokens of the hidden representation of the input sequence $H^L$. In the pooling-based shortening the sequence (of size $M$) is divided into non-overlapping groups of $K$ neighboring tokens each ($K$ is a hyper-parameter). Pooling of the tokens in each group is then performed:
\begin{equation} \label{eq:shortening-pooling}
\begin{aligned}
&\tilde{G} = \pooling( H^L ),
\end{aligned}
\end{equation}
where $\tilde{G}$ is the sequence of size $\lceil M/K \rceil$ of the pooled tokens. Subsequently, the pooled tokens $\tilde{G}$ attend to the hidden representation of the original sequence using the attention module followed by the residual connection and the Layer Normalization:
\begin{equation} \label{eq:shortening-attention}
\begin{aligned}
&G = \layernorm(\tilde{G} + \attn(\tilde{G}, H^L, H^L)),
\end{aligned}
\end{equation}
where $G$ is the final shortened sequence. Commonly used pooling operations are average \citep{dai2020funnel} and linear pooling \citep{nawrot-etal-2022-hierarchical} (learned linear transformation of the concatenated tokens).

\section{Method}

\subsection{Latent Grouping}
\label{sec:grouping}

\begin{figure}[!ht]
    \center{}
    \includegraphics[width=0.57\linewidth]{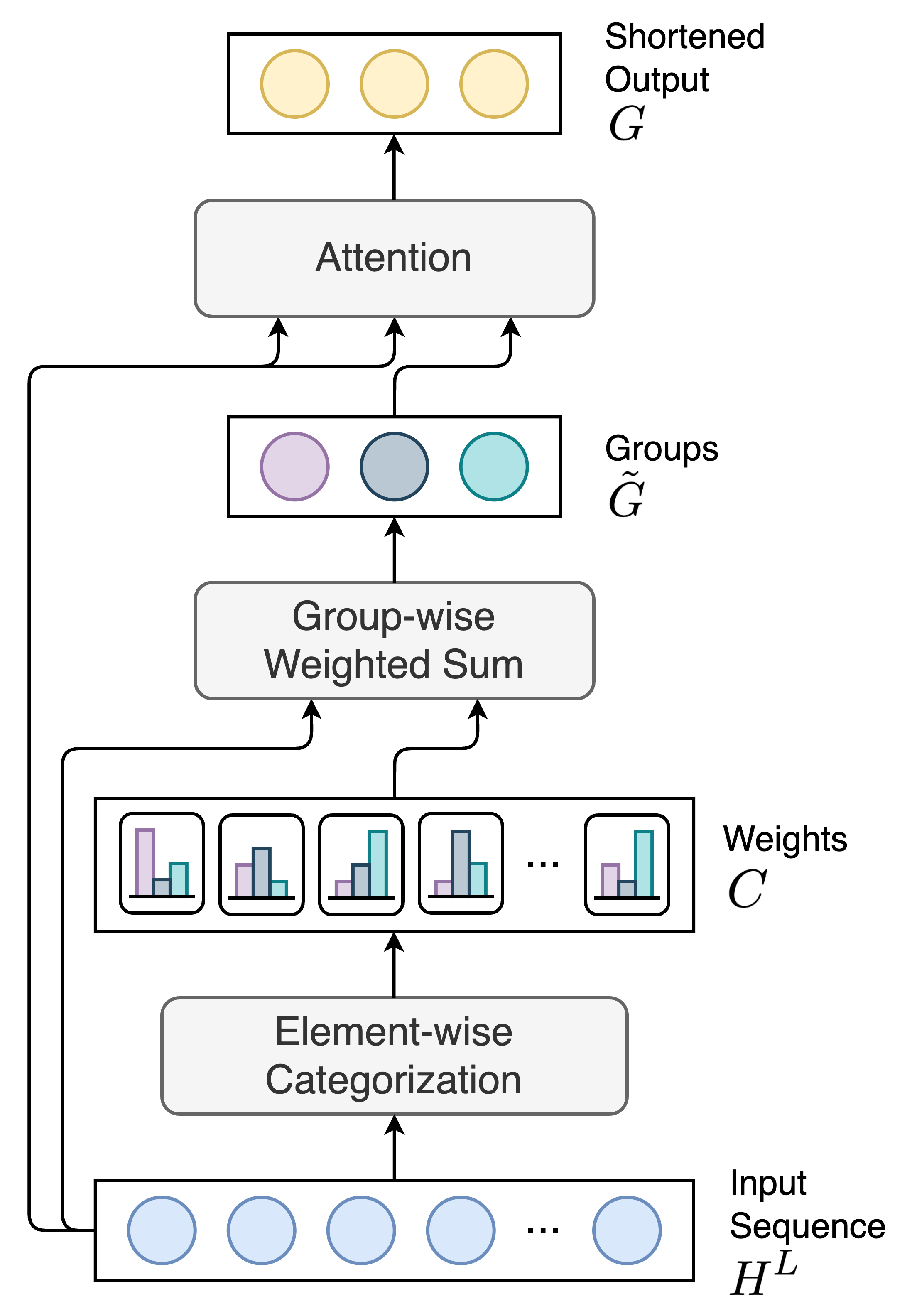}
    \caption{Illustration of Latent Grouping shortening with the number of groups set to three.}
    \label{fig:grouping-shortening}
\end{figure} 

In contrast to pooling, Latent Grouping, illustrated in Figure~\ref{fig:grouping-shortening}, results in a fixed number of tokens in the shortened sequence corresponding to the number of groups $K$, which is a hyper-parameter. Each token is categorized into a group by the feed-forward network with the number of outputs equal to the number of groups. We obtain the categorization for the $i$-th token to $k$-th group $c_{i,k}$ by applying the Softmax function to the outputs in the dimension of the groups:
\begin{equation} \label{eq:grouping-categorize}
\begin{aligned}
&\mathbf{c}_{i} = \softmax( \ffn( \mathbf{h}^L_i ) ), \\
&\forall i = 1, ..., M,
\end{aligned}
\end{equation}
where $\mathbf{h}^L$ is the hidden representation of the last encoder layer and $\mathbf{c}_i$ is the vector of size $K$ representing the categorizations of the $i$-th token to all the groups. As an alternative to Softmax, Sparsemax function \citep{martins2016softmax} can also be used resulting in the categorizations of tokens that are more sparse, which means that a token is categorized into a smaller number of groups, and most categorizations are equal to zero. Subsequently, the groups $\tilde{G}$ are constructed as the sum of the hidden representations $\mathbf{h}^L$ with categorizations $c_{i, k}$ used as weights:
\begin{equation} \label{eq:grouping-sum}
\begin{aligned}
&\tilde{\mathbf{g}}_k = \sum_{i} c_{i,k} \mathbf{h}^L_i , \\ 
&\forall k = 1, ..., K,
\end{aligned}
\end{equation}
where $\tilde{\mathbf{g}}_k$ is a $k$-th grouped token composing the sequence $\tilde{G}$ in the equation~(\ref{eq:shortening-pooling}). The network learns how to soft-assign each token to the groups. A group representation is computed using the weighted average of tokens, which makes backpropagation into the categorizing network possible. Finally, the attention module is applied as in equation~(\ref{eq:shortening-attention}).

\begin{figure*}[!ht]
    \centering
    \includegraphics[width=0.8\linewidth]{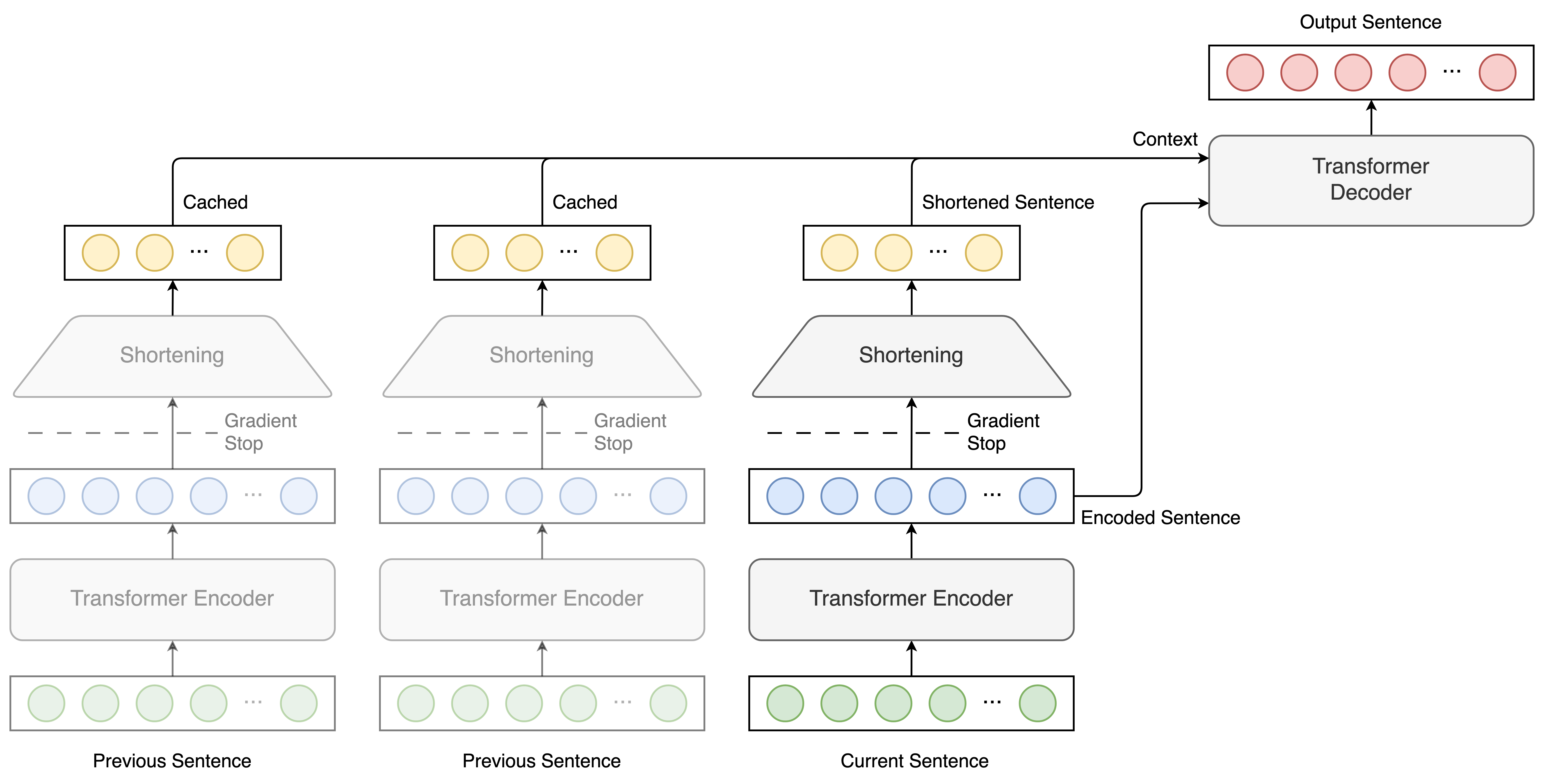}
    \caption{The illustration of a Shortening Architecture with the representation of the two previous sentences being cached. The dashed line represents the optional blocking of the gradient during training.}
    \label{fig:shortening-architecture}
\end{figure*}

\subsection{Latent Selecting}
\label{sec:selecting}

Latent Selecting differs from Latent Grouping by enabling the groups to select tokens to aggregate rather than assigning each token to a group and allowing the model to ignore tokens entirely rather than assigning them to at least one group. This is similar to selecting the \textit{hub} tokens in Power-BERT \cite{goyal2020power}, where the selection is based on the attention scores of the previous layer. Although Latent Selecting can be achieved by maintaining a categorizing feed-forward network for each group, we utilize the same network as described for Latent Grouping but apply the Softmax (or Sparsemax) function in equation~(\ref{eq:grouping-categorize}) in the sequence dimension instead of the token dimension. 

\subsection{Context Shortening}

The architecture we use, illustrated in Figure~\ref{fig:shortening-architecture}, is based on caching the hidden representations produced by the encoder, where the representations of the tokens of the current sentence are stored and can be reused as context when the subsequent sentences are translated. Although this architecture uses only a single encoder, it is different from the single-encoder models because the current sentence and the context sentences are processed separately. While in the standard caching architecture the hidden representation of all the tokens is stored, we introduce a Sequence Shortening module directly after the encoder, which returns the compressed hidden representation usually containing fewer tokens than the original sequence. We consider: mean pooling \cite{dai2020funnel}, max pooling, linear pooling \cite{nawrot-etal-2022-hierarchical}, Latent Grouping, and Latent Selecting. Additionally, we also test the simple aggregation of the whole context sequences into a single vector by averaging the tokens. Conceptually, Sequence Shortening of the context can be seen as a middle-ground between storing tokens and sentence aggregations. 

The integration of the context with the decoder can also be done in several ways. Firstly, the context sentences can be concatenated to the current sentence. This method is similar to the single-encoder (concatenating) architecture, where the difference is that the encoder does not have access to other sentences in the case of caching architecture. In this case, the decoder layers are the same as in the vanilla transformer with the self- and cross-attention modules. Secondly, the context sentences can be processed in the decoder layers by a separate context-attention module, where the decoder tokens attend to the context tokens. We experiment with the parallel and serial alignment of the cross- and context-attention modules. Additionally, we also experiment with gating the representation resulting from applying context-attention using the following equation:
\begin{equation} \label{eq:context-attention-gate}
\begin{aligned}
&\lambda_i = \sigma(\ffn(\mathbf{\hat{h}}_i)), \\ 
&\mathbf{\hat{h}}'_i = \lambda_i \mathbf{\hat{h}}_i, \\ 
&\forall i = 1, ..., M
\end{aligned}
\end{equation}
where $\mathbf{\hat{h}}_i$ is the $i$-th token representation returned by the context-attention module, $\ffn$ is a token-wise linear layer with one output, $\sigma$ is the Sigmoid function.

For Sentence Aggregation and Shortening architectures, the aggregated or shortened representation of the current sentence can be included in context sentences. This helps with the training, as often none of the previous sentences has an effect on the translation, known as the two-fold sparsity problem \citep{lupo-etal-2022-divide}, and the context attention module can still be trained to attend to the representation of the current sentence. To allow the decoder to distinguish between context sentences we employ learned segment embeddings \citep{devlin-etal-2019-bert}. Similarly, we also add learned positional encoding for the shortened tokens inside context sentences.

During training, caching is not used, meaning that the model receives tokenized context sentences and processes them using the same encoder. This implies that the weights of the encoder receive the backpropagated error from multiple sources - the current sentence and each of the context sentences, which can lead to difficulties in training. Therefore, we consider blocking the gradient after the encoder and before shortening (where applicable) by allowing the gradient information to flow for a specified number of context sentences, after which, the gradient is blocked.

\section{Experiments}
\label{sec:experiments}

All our experiments are implemented\footnote{The code for this paper (based on \url{https://github.com/neulab/contextual-mt}) can be found on Github \url{https://github.com/Pawel-M/shortening-context-mt}.} in \textit{fairseq} framework \citep{ott-etal-2019-fairseq}. We used the code repository of \citet{fernandes-etal-2021-measuring} as the base for our implementation. 

\subsection{Data}

\begin{table}[!ht]
    \centering
    \begin{tabular}{lrrrrrr}
    \hline
        \textbf{Dataset} & \textbf{Docs} & \textbf{Sent/Doc} & \textbf{Tok/Sent} \\ 
        \hline
        En-De Train & 1698 & 121.4 & 21.9 \\ 
        En-De Valid & 62 & 87.6 & 20.6 \\
        En-De Test & 12 & 90.0 & 20.8 \\ 
        \hline
        En-Fr Train  & 1914 & 121.6 & 22.0 \\ 
        En-Fr Valid  & 66 & 88.2 & 20.9 \\ 
        En-Fr Test  & 12 & 100.8 & 21.4  \\ 
    \hline
    \end{tabular}
    \caption{The details of the IWSLT 2017 datasets.}
    \label{tab:dataset-stats}
\end{table}

\begin{table*}[!ht]
    \centering
    \begin{tabular}{lrrrrrr}
    \hline
        \textbf{Model} & \textbf{BLEU} & \textbf{Accuracy} & \textbf{} & \textbf{} & \textbf{}   \\ \hline
        Sentence-level & 28.11 & 43.67\% & ~ & ~ & ~ & ~   \\ 
        \hline
        ~ & \multicolumn{2}{c}{\textbf{Context: 1}} & \multicolumn{2}{c}{\textbf{Context: 2}} & \multicolumn{2}{c}{\textbf{Context: 3}}   \\ 
        \textbf{Model} & \textbf{BLEU} & \textbf{Accuracy} & \textbf{BLEU} & \textbf{Accuracy} & \textbf{BLEU} & \textbf{Accuracy} \\ 
        \hline
        Single-encoder  & 28.31 & 47.42\% & 27.95 & 48.18\% & 27.88 & 48.88\%   \\ 
        Multi-encoder & \textbf{28.67} & 44.93\% & 28.50 & 46.65\% & 28.26 & 45.00\%   \\ 
        \hline
        Caching Tokens & 28.35 & 54.06\% & 28.50 & 54.13\% & \textbf{29.08} & 51.23\%  \\ 
        Caching Sentence & 28.38  & 45.72\% & 26.73 & 45.26\% & 26.70 & 44.91\%  \\ 
        Shortening - Max Pooling & 27.62 & 51.67\% & 27.88 & \textbf{55.08\%} & 28.26 & 50.89\%  \\ 
        Shortening - Avg Pooling & 28.09 & 53.37\% & 27.85 & 54.81\% & 28.38 & 50.54\%  \\ 
        Shortening - Linear Pooling & 27.62 & 52.71\% & 28.03 & 52.13\% & 28.18 & 51.27\%  \\ 
        Shortening - Grouping & 28.21 & \textbf{56.98\%} & \textbf{28.70} & 54.51\% & 28.49 & 51.16\%  \\ 
        Shortening - Selecting & 28.15 & 54.48\% & 28.55 & 54.21\% & 28.01 & \textbf{51.95\%} \\
        \hline
    \end{tabular}
    \caption{Results of the \textbf{En-De} IWSLT 2017 experiment. The models were trained to use only the source-side context. We report BLEU of the test subset and the accuracy of the ContraPro \citep{muller-etal-2018-large} contrastive dataset.}
    \label{tab:results-iwslt-en-de}
\end{table*}

We used the English-to-German and English-to-French directions of the IWSLT 2017 \citep{cettolo-etal-2017-overview} document-level dataset that is based on the subtitles of the TED Talks\footnote{\url{https://www.ted.com/}}. Following \citet{fernandes-etal-2021-measuring}, we used \textit{tst2011}-\textit{tst2014} as validation subset and \textit{tst2015} as the test subset. The data is byte-pair encoded \citep{sennrich-etal-2016-neural} using SentencePiece framework \citep{kudo-richardson-2018-sentencepiece} on the training subset with 20,000 vocabulary size for each language separately (see Table~\ref{tab:dataset-stats}). We measured  BLEU \citep{papineni-etal-2002-bleu} using \textit{sacreBleu} library \cite{post-2018-call}. We also report COMET \citep{rei-etal-2020-comet} in Appendix~\ref{sec:additional-result}.

To measure the context usage of the trained models, we employed ContraPro \citep{muller-etal-2018-large} contrastive dataset for the English-to-German direction, and the contrastive dataset by \citet{lopes-etal-2020-document} for the English-to-French direction. Both are based on the OpenSubtitles 2018 dataset \citep{lison-etal-2018-opensubtitles2018}. These datasets consist of the source sentence with the context (previous sentences on the source and target side) with several translations differing only in a pronoun that requires context to be correctly translated. Models rank the translations by assigning probabilities to each of them. The translation is considered to be accurate when the right translation is ranked the highest by the model.

\subsection{Models}

Based on the described methods, we trained the following caching models:
\begin{itemize}[topsep=0pt,itemsep=0pt,partopsep=0pt, parsep=0pt]
    \item \textbf{Caching Tokens} - where the encoder representations of the context sentences are stored directly,
    \item \textbf{Caching Sentence} - where the representations of the context sentences are averaged and stored, 
    \item \textbf{Shortening - Avg Pooling} - Sequence shortening with mean pooling applied to the outputs of the encoder, based on \citep{dai2020funnel},
    \item \textbf{Shortening - Max Pooling} - shortening with max pooling,
    \item \textbf{Shortening - Linear Pooling} - shortening with linear pooling, based on \citep{nawrot-etal-2022-hierarchical},
    \item \textbf{Shortening - Grouping} - shortening with Latent Grouping (Section~\ref{sec:grouping}),
    \item \textbf{Shortening - Selecting} - shortening with Latent Selecting (Section~\ref{sec:selecting}).
\end{itemize}
For all the aggregating models, the current sentence is also used as context and is concatenated with the context sentences after embedding.
Moreover, we also test the following baseline models:
\begin{itemize}[topsep=0pt,itemsep=0pt,partopsep=0pt, parsep=0pt]
    \item \textbf{Sentence-level Transformer} - where context sentences are ignored,
    \item \textbf{Single-encoder Transformer} - where context sentences are prepended to the current sentence and processed by the encoder, we used \citet{fernandes-etal-2021-measuring} implementation,
    \item \textbf{Multi-encoder Transformer} - with the separate encoder (without weights-sharing) used to encode the context sentences, again based on the \citet{fernandes-etal-2021-measuring} implementation, where the context and the current sentence are concatenated in the decoder. Our experiments revealed that this integration yields better results than with the separate context-attention module.
\end{itemize}

All tested models are based on the Transformer base architecture \citep{vaswani2017attention}. The hyper-parameters and model details can be found in Appendix~\ref{sec:models-and-training-details}. We tuned the hyper-parameters of the models based on the performance on the validation subset. From the K values of $[2, 3, 4]$ for pooling architectures $2$ was selected. For grouping and selecting architectures, we considered K values of $[8, 9, 10, 11]$ and selected $9$ and $10$ respectively for he English-to-German direction and $11$ (for both models) for the English-to-French direction. For the categorizing network, we used one hidden layer with $512$ units and the Sparsemax activation function to obtain more sparse categorizations in an effort to increase the interpretability of the models \citep{correia-etal-2019-adaptively, meister-etal-2021-sparse}. We performed preliminary experiments to find the architectural choices (gradient stopping and the decoder integration) for each caching model. In Caching Tokens, Caching Sentence, and Pooling architectures, we block gradient past the encoder for context sentences. Additionally, we allow gradient into the shortening from one and two context sentences for Selecting and Grouping architectures respectively. All models apart from Caching Sentence use sequential attention modules in the decoder (self-attention, cross-attention, and context-attention) without any gating mechanism. Caching Sentence yields the highest performance when parallel cross- and context-attention decoder is used with the gate on the context branch (see equation~(\ref{eq:context-attention-gate})).

\subsection{Results}

\begin{table*}[!ht]
    \centering
    \begin{tabular}{lrrrrrr}
    \hline
        \textbf{Model} & \textbf{BLEU} & \textbf{Accuracy} & \textbf{} & \textbf{} & \textbf{}   \\ \hline
        Sentence-level & 37.64 & 75.92\% & ~ & ~ & ~ & ~   \\ 
        \hline
        ~ & \multicolumn{2}{c}{\textbf{Context: 1}} & \multicolumn{2}{c}{\textbf{Context: 2}} & \multicolumn{2}{c}{\textbf{Context: 3}}   \\ 
        \textbf{Model} & \textbf{BLEU} & \textbf{Accuracy} & \textbf{BLEU} & \textbf{Accuracy} & \textbf{BLEU} & \textbf{Accuracy} \\ 
        \hline
        Single-encoder  & 37.25 & 77.27\% & 37.18 & 78.98\% & 37.12 & \textbf{80.87\%}   \\ 
        Multi-encoder & 37.44 & 75.72\% & 37.12 & 77.23\% & 37.34 & 75.76\%   \\ 
        \hline
        Caching Tokens & 36.88 & 79.67\% & 37.29 & 80.14\% & 37.73 & 79.90\%  \\ 
        Caching Sentence & 36.50 & 77.33\% & 34.21 & 76.25\% & 34.78 & 75.71\%  \\ 
        Shortening - Max Pooling & \textbf{37.48} & 79.51\% & 36.72 & 80.59\% & 37.85 & 79.71\%  \\ 
        Shortening - Avg Pooling & 37.13 & 77.75\% & 37.12 & 80.16\% & \textbf{38.18} & 80.41\%  \\ 
        Shortening - Linear Pooling & 37.02 & 80.47\% & 37.12 & 79.37\% & 37.42 & 79.64\%  \\ 
        Shortening - Grouping & 37.05 & 79.91\% & \textbf{37.98} & \textbf{81.13\%} & 37.18 & 79.54\%  \\ 
        Shortening - Selecting & 37.38 & \textbf{80.89\%} & 37.83 & 80.32\% & 37.81 & 80.09\% \\
        \hline
    \end{tabular}
    \caption{Results of the \textbf{En-Fr} IWSLT 2017 experiment. The models were trained to use only the source-side context. We report BLEU of the test subset and the accuracy of the contrastive dataset by \citet{lopes-etal-2020-document}.}
    \label{tab:results-iwslt-en-fr}
\end{table*}

The results of the single run (with the predetermined seed) of the English-to-German translation on the IWSLT 2017 dataset up to the context size of three can be seen in Table~\ref{tab:results-iwslt-en-de}. The BLEU score of the context-aware models is generally similar to or slightly higher than the sentence-level Transformer. BLEU does not correlate well with the contrastive accuracy, which is strictly higher for all context-aware models. This confirms that sentence-level metrics do not reflect the context usage of the models. 
The highest contrastive dataset accuracy was achieved by the Grouping Shortening model for the context size of one, the Max Pooling Shortening model for the context size of two, and the Selecting Shortening model for the context size of three. 
The highest accuracy averaged over the context sizes up to three was reached by the model employing Latent Grouping, followed by the Latent Selecting model. 
Caching Tokens architecture exhibits comparable BLEU scores to the Single- and Multi-encoder architectures while achieving higher accuracy on the contrastive dataset. Caching Sentence architecture performed worse than other tested models, suggesting that representing the whole sentence as a single vector is not sufficient for contextual translation.

Table~\ref{tab:results-iwslt-en-fr} shows the results of the English-to-French translation with the context size up to three. The BLEU scores of all models are comparable (apart from the Caching Sentence architecture). Latent Grouping achieved the highest accuracy on the contrastive dataset for the context size of one, and Latent Selecting and Single-encoder architectures for the context sizes of one and three, respectively. The results in terms of COMET \citep{rei-etal-2020-comet} can be found in Appendix~\ref{sec:additional-result}. The detailed results of the performance of the models on the contrastive datasets are presented in Appendix~\ref{sec:detailed-contrastive-result}.We show several examples of translations by the tested models in Appendix~\ref{sec:examples-results}.

\begin{figure}[!ht]
\center{}
    \includegraphics[width=1\linewidth]{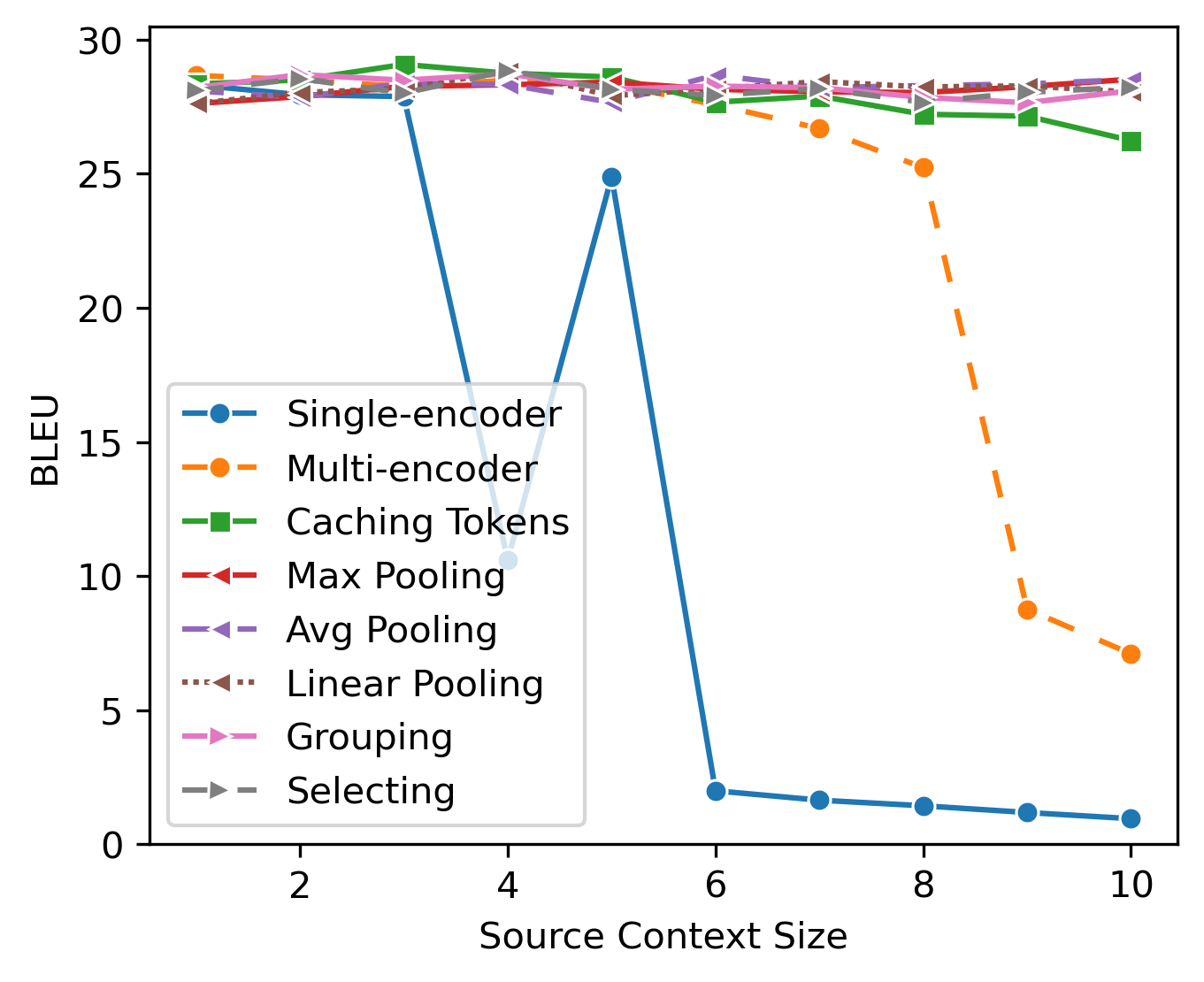}
    \caption{BLEU of the models trained on the \textbf{En-De} IWSLT 2017 dataset with the context sizes up to 10. Caching Sentence model was not included for clarity. }
    \label{fig:extended-context}
\end{figure}

Caching Tokens and Shortening models achieved higher accuracies than the Single- and Multi-encoder architectures (with the exception of Single-encoder on the English-to-French translation with the context size of three). In order to examine the effectiveness of the investigated architectures on even longer contexts we trained the models on the English-to-German IWSLT 2017 dataset with context sizes of up to 10. The results in terms of BLEU can be seen in Figure~\ref{fig:extended-context}. The detailed results (in terms of BLEU, COMET, and the accuracy on the ContraPro dataset) are presented in Appendix~\ref{sec:larger-context-result}. The performance of the models employing Sequence Shortening is relatively high and stable for all tested context sizes. The caching architecture shows the reduction in BLEU for context sizes of 8 to 10 compared to the shortening architectures. We attribute the poor performance of the single-encoder (and to an extent multi-encoder) architecture to the large input sizes and the small size of the training dataset. 

Applying Sequence Shortening to the cached sentence does not hurt the performance and exhibits more stable training with the long context sizes while reducing the memory footprint of the inference (Section~\ref{sec:memory-analysis}). Furthermore, Latent Grouping and Latent Selecting are increasing the interpretability of the model through the sparse assignment of tokens into groups (Section~\ref{sec:grouping-visualization}).

\begin{figure}[!ht]
\center{}
    \begin{subfigure}{0.8\linewidth}
        \includegraphics[width=1\linewidth]{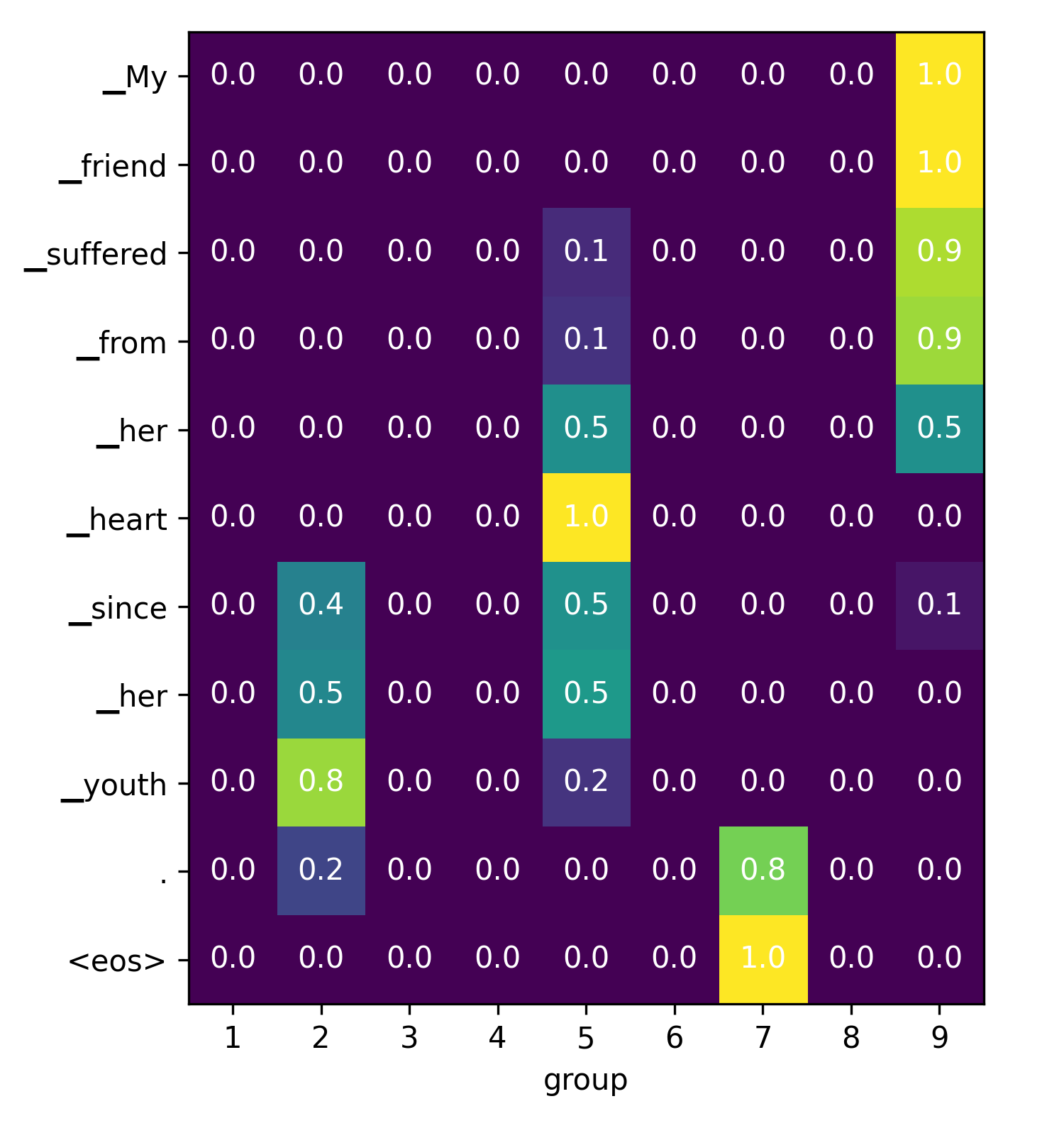}
        \caption{Latent Grouping}
        \label{fig:visualization-en-de-8-grouping}
    \end{subfigure}\hfill
    \begin{subfigure}{0.83\linewidth}
        \includegraphics[width=1\linewidth]{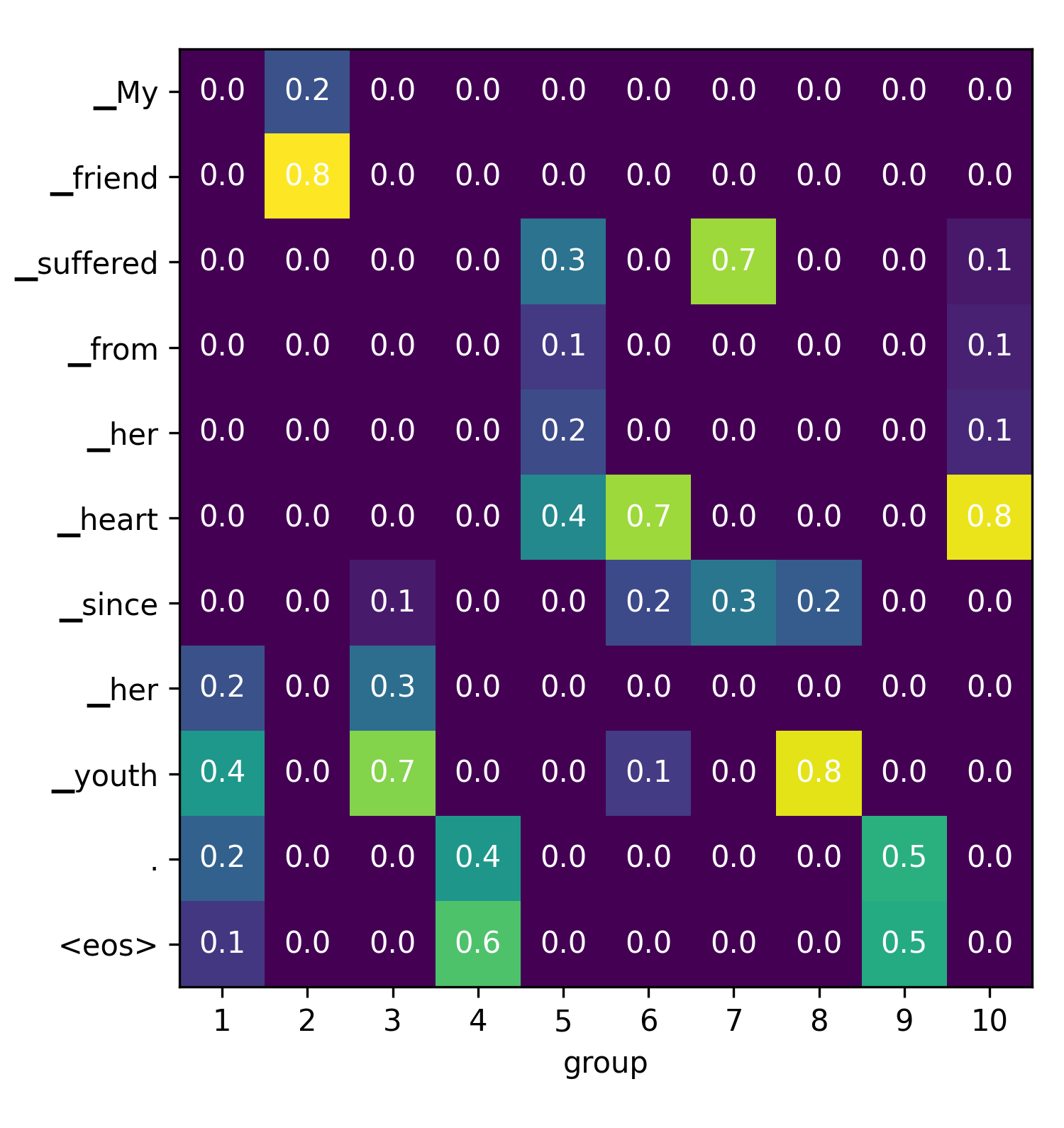}
        \caption{Latent Selecting}
        \label{fig:visualization-en-de-8-selecting}
    \end{subfigure}\hfill
    
    \caption{Visualization of tokens of the sentence from the ContraPro dataset grouped (\ref{fig:visualization-en-de-8-grouping}) and selected (\ref{fig:visualization-en-de-8-selecting}) by the model using Latent Grouping and Latent Selecting.}
    \label{fig:visualization-en-de-8}
\end{figure}

\subsection{Token Assignment Visualization}
\label{sec:grouping-visualization}

An example visualization of groupings and selections of the Latent Grouping and Selecting architectures can be seen in Figure~\ref{fig:visualization-en-de-8} and more can be found in Appendix~\ref{sec:exteded-visualizaitons}. Latent Grouping seems to group tokens according to position with nouns given a high categorization score within a group. Furthermore, some groups contain more tokens than other groups. We hypothesize that the groups that contain more tokens are responsible for the general sense of the sentence and the groups with less tokens are responsible for encoding the details. Surprisingly, only four groups out of nine are utilized by the model. We hypothesize that the rest are used as the \textit{no-op} tokens \citep{clark-etal-2019-bert} in the context-attention when the context is not needed. Latent Selecting, by design, has to assign tokens to each group. Again, nouns seem to be included in a group more often than other parts of speech. Some groups select punctuation marks and the \texttt{<eos>} token, which could take the role of the \textit{no-op} tokens. 

\subsection{Memory Usage}
\label{sec:memory-analysis}

We measured the memory used by the tested models as the value returned by the \verb|torch.cuda.max_memory_allocated()| function. For clarity we omit the Caching Sentence model (as the worst performing) and the Max Pooling model (with results the same as the Avg Pooling model). We report the operation memory - the memory during inference on top of the memory taken by the model itself - on the examples from the test subset of the English-to-German IWSLT 2017 dataset with different numbers of context sentences. For context sizes above three, we used the models trained on the context size of three in order to not disadvantage the Single- and Multi-encoder architectures that were not able to learn on the dataset for large context sizes. The results are presented in Figure~\ref{fig:memory-analysis}. Although the number of parameters (see Appendix~\ref{sec:models-and-training-details}) is a dominant factor determining the overall memory usage, the operation memory grows at different paces for different architectures with the increased context size. The operational memory of the Single- and Multi-encoder models grows quadratically, while for caching and shortening architectures it grows linearly. Furthermore, the rate of increase is slower for shortening architectures compared to the Caching Tokens architecture, which can allow the significant advantage of shortening in the setting of long sentences or large contexts.

\begin{figure}[!ht]
\center{}
    \includegraphics[width=1\linewidth]{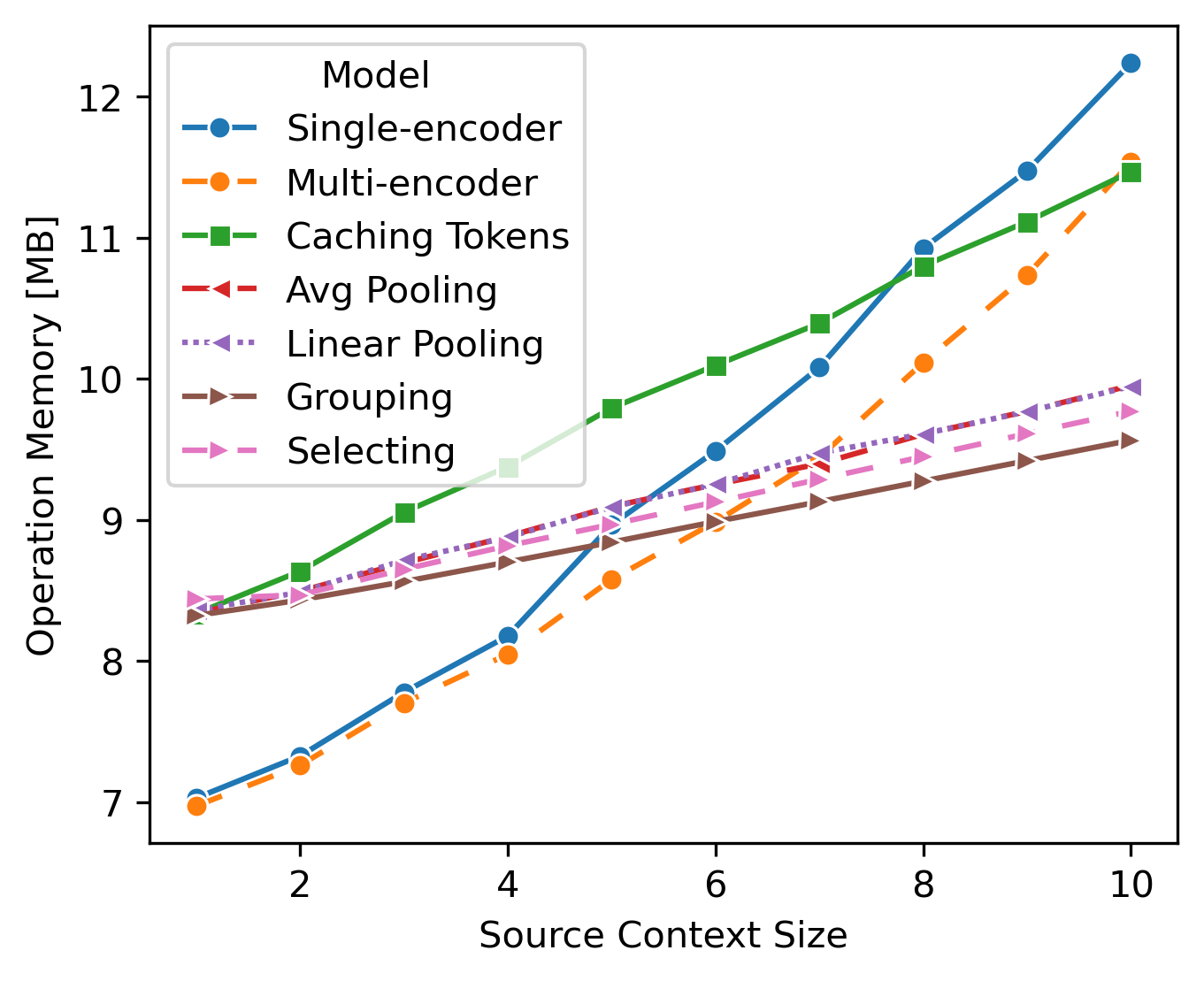}
    \caption{The mean operation memory of the models when performing inference on the examples from the \textbf{En-De} IWSLT 2017 test subset with the varying context sizes. For the context sizes above three, we used the models trained on the context size of three.}
    \label{fig:memory-analysis}
\end{figure}

\section{Conclusions}

Caching architectures for Context-aware Machine Translation have not been widely explored in the literature so far. In this study, we show that a simple method of remembering the hidden representations of the previous sentences is comparable with more established Single- and Multi-encoder approaches in terms of BLEU and can be more effective in capturing context (up to 6 percentage points of the accuracy on the contrastive dataset for the context size of one) in the relatively low-resource training scenario. Furthermore, the caching architectures are more stable to train in the regime of larger context sizes according to our experiments.

Pooling-based shortening of the cached sentence maintains the comparable results to the caching architecture, while our introduced shortening methods - Latent Grouping and Selecting - show on average a strong performance both in terms of BLEU and accuracy while maintaining slower growth of the memory usage during inference, and potential increased interpretability of the model through sparse assignment of tokens into groups. Sequence Shortening, in general, exhibit stable training in the regime of large context sizes compared to other tested methods. In future work, we will explore the integration of Sequence Shortening with the target-side context.

\section{Limitations}

Our investigation is limited to the source-side context. There exist linguistic phenomena that can only be addressed by using target-side context \cite{voita-etal-2019-good}. While both caching and shortening could be applied to the target side as well, we do not provide an empirical evaluation of the performance of this approach.

Additionally, we do not apply sentence-level pre-training to our models. Architectures using Sequence Shortening could benefit from multiple stages of pre-training.

Lastly, our experiments involve language pairs from the same language family (English-to-German and English-to-French). We trained the models using the relatively low-resource datasets (IWSLT 2017) and the contrastive datasets used in this work target only the pronoun disambiguation task.

\section{Acknowledgments}
The research presented in this paper was conducted as part of VOXReality project\footnote{\url{https://voxreality.eu/}}, which was funded by the European Union Horizon Europe program under grant agreement No 101070521.

\bibliography{anthology,custom}

\begin{thebibliography}{66}
\expandafter\ifx\csname natexlab\endcsname\relax\def\natexlab#1{#1}\fi

\bibitem[{Agrawal et~al.(2018)Agrawal, Turchi, and
  Negri}]{agrawal2018contextual}
Ruchit Agrawal, Marco Turchi, and Matteo Negri. 2018.
\newblock Contextual handling in neural machine translation: Look behind, ahead
  and on both sides.
\newblock In \emph{Proceedings of the 21st Annual Conference of the European
  Association for Machine Translation}, pages 31--40.

\bibitem[{Ba et~al.(2016)Ba, Kiros, and Hinton}]{ba2016layer}
Jimmy~Lei Ba, Jamie~Ryan Kiros, and Geoffrey~E Hinton. 2016.
\newblock Layer normalization.
\newblock \emph{arXiv preprint arXiv:1607.06450}.

\bibitem[{Bao et~al.(2021)Bao, Zhang, Teng, Chen, and Luo}]{bao-etal-2021-g}
Guangsheng Bao, Yue Zhang, Zhiyang Teng, Boxing Chen, and Weihua Luo. 2021.
\newblock \href {https://doi.org/10.18653/v1/2021.acl-long.267}
  {{G}-transformer for document-level machine translation}.
\newblock In \emph{Proceedings of the 59th Annual Meeting of the Association
  for Computational Linguistics and the 11th International Joint Conference on
  Natural Language Processing (Volume 1: Long Papers)}, pages 3442--3455,
  Online. Association for Computational Linguistics.

\bibitem[{Bawden et~al.(2018)Bawden, Sennrich, Birch, and
  Haddow}]{bawden-etal-2018-evaluating}
Rachel Bawden, Rico Sennrich, Alexandra Birch, and Barry Haddow. 2018.
\newblock \href {https://doi.org/10.18653/v1/N18-1118} {Evaluating discourse
  phenomena in neural machine translation}.
\newblock In \emph{Proceedings of the 2018 Conference of the North {A}merican
  Chapter of the Association for Computational Linguistics: Human Language
  Technologies, Volume 1 (Long Papers)}, pages 1304--1313, New Orleans,
  Louisiana. Association for Computational Linguistics.

\bibitem[{Beltagy et~al.(2020)Beltagy, Peters, and
  Cohan}]{beltagy2020longformer}
Iz~Beltagy, Matthew~E Peters, and Arman Cohan. 2020.
\newblock Longformer: The long-document transformer.
\newblock \emph{arXiv preprint arXiv:2004.05150}.

\bibitem[{Bulatov et~al.(2022)Bulatov, Kuratov, and
  Burtsev}]{bulatov2022recurrent}
Aydar Bulatov, Yury Kuratov, and Mikhail Burtsev. 2022.
\newblock Recurrent memory transformer.
\newblock \emph{Advances in Neural Information Processing Systems},
  35:11079--11091.

\bibitem[{Cettolo et~al.(2017)Cettolo, Federico, Bentivogli, Niehues,
  St{\"u}ker, Sudoh, Yoshino, and Federmann}]{cettolo-etal-2017-overview}
Mauro Cettolo, Marcello Federico, Luisa Bentivogli, Jan Niehues, Sebastian
  St{\"u}ker, Katsuhito Sudoh, Koichiro Yoshino, and Christian Federmann. 2017.
\newblock \href {https://aclanthology.org/2017.iwslt-1.1} {Overview of the
  {IWSLT} 2017 evaluation campaign}.
\newblock In \emph{Proceedings of the 14th International Conference on Spoken
  Language Translation}, pages 2--14, Tokyo, Japan. International Workshop on
  Spoken Language Translation.

\bibitem[{Chen et~al.(2022)Chen, Li, Gong, Zhang, and Zhou}]{chen2022one}
Linqing Chen, Junhui Li, Zhengxian Gong, Min Zhang, and Guodong Zhou. 2022.
\newblock \href {https://doi.org/10.1145/3526215} {One type context is not
  enough: Global context-aware neural machine translation}.
\newblock \emph{ACM Trans. Asian Low-Resour. Lang. Inf. Process.}, 21(6).

\bibitem[{Clark et~al.(2019)Clark, Khandelwal, Levy, and
  Manning}]{clark-etal-2019-bert}
Kevin Clark, Urvashi Khandelwal, Omer Levy, and Christopher~D. Manning. 2019.
\newblock \href {https://doi.org/10.18653/v1/W19-4828} {What does {BERT} look
  at? an analysis of {BERT}{'}s attention}.
\newblock In \emph{Proceedings of the 2019 ACL Workshop BlackboxNLP: Analyzing
  and Interpreting Neural Networks for NLP}, pages 276--286, Florence, Italy.
  Association for Computational Linguistics.

\bibitem[{Correia et~al.(2019)Correia, Niculae, and
  Martins}]{correia-etal-2019-adaptively}
Gon{\c{c}}alo~M. Correia, Vlad Niculae, and Andr{\'e} F.~T. Martins. 2019.
\newblock \href {https://doi.org/10.18653/v1/D19-1223} {Adaptively sparse
  transformers}.
\newblock In \emph{Proceedings of the 2019 Conference on Empirical Methods in
  Natural Language Processing and the 9th International Joint Conference on
  Natural Language Processing (EMNLP-IJCNLP)}, pages 2174--2184, Hong Kong,
  China. Association for Computational Linguistics.

\bibitem[{Costa-juss{\`a} et~al.(2022)Costa-juss{\`a}, Cross, {\c{C}}elebi,
  Elbayad, Heafield, Heffernan, Kalbassi, Lam, Licht, Maillard
  et~al.}]{costa2022no}
Marta~R Costa-juss{\`a}, James Cross, Onur {\c{C}}elebi, Maha Elbayad, Kenneth
  Heafield, Kevin Heffernan, Elahe Kalbassi, Janice Lam, Daniel Licht, Jean
  Maillard, et~al. 2022.
\newblock No language left behind: Scaling human-centered machine translation.
\newblock \emph{arXiv preprint arXiv:2207.04672}.

\bibitem[{Dai et~al.(2020)Dai, Lai, Yang, and Le}]{dai2020funnel}
Zihang Dai, Guokun Lai, Yiming Yang, and Quoc Le. 2020.
\newblock Funnel-transformer: Filtering out sequential redundancy for efficient
  language processing.
\newblock \emph{Advances in neural information processing systems},
  33:4271--4282.

\bibitem[{Dai et~al.(2019)Dai, Yang, Yang, Carbonell, Le, and
  Salakhutdinov}]{dai-etal-2019-transformer}
Zihang Dai, Zhilin Yang, Yiming Yang, Jaime Carbonell, Quoc Le, and Ruslan
  Salakhutdinov. 2019.
\newblock \href {https://doi.org/10.18653/v1/P19-1285} {Transformer-{XL}:
  Attentive language models beyond a fixed-length context}.
\newblock In \emph{Proceedings of the 57th Annual Meeting of the Association
  for Computational Linguistics}, pages 2978--2988, Florence, Italy.
  Association for Computational Linguistics.

\bibitem[{Devlin et~al.(2019)Devlin, Chang, Lee, and
  Toutanova}]{devlin-etal-2019-bert}
Jacob Devlin, Ming-Wei Chang, Kenton Lee, and Kristina Toutanova. 2019.
\newblock \href {https://doi.org/10.18653/v1/N19-1423} {{BERT}: Pre-training of
  deep bidirectional transformers for language understanding}.
\newblock In \emph{Proceedings of the 2019 Conference of the North {A}merican
  Chapter of the Association for Computational Linguistics: Human Language
  Technologies, Volume 1 (Long and Short Papers)}, pages 4171--4186,
  Minneapolis, Minnesota. Association for Computational Linguistics.

\bibitem[{Feng et~al.(2022)Feng, Li, Song, Zheng, and
  Koehn}]{feng-etal-2022-learn}
Yukun Feng, Feng Li, Ziang Song, Boyuan Zheng, and Philipp Koehn. 2022.
\newblock \href {https://doi.org/10.18653/v1/2022.findings-naacl.105} {Learn to
  remember: Transformer with recurrent memory for document-level machine
  translation}.
\newblock In \emph{Findings of the Association for Computational Linguistics:
  NAACL 2022}, pages 1409--1420, Seattle, United States. Association for
  Computational Linguistics.

\bibitem[{Fernandes et~al.(2023)Fernandes, Yin, Liu, Martins, and
  Neubig}]{fernandes-etal-2023-translation}
Patrick Fernandes, Kayo Yin, Emmy Liu, Andr{\'e} Martins, and Graham Neubig.
  2023.
\newblock \href {https://doi.org/10.18653/v1/2023.acl-long.36} {When does
  translation require context? a data-driven, multilingual exploration}.
\newblock In \emph{Proceedings of the 61st Annual Meeting of the Association
  for Computational Linguistics (Volume 1: Long Papers)}, pages 606--626,
  Toronto, Canada. Association for Computational Linguistics.

\bibitem[{Fernandes et~al.(2021)Fernandes, Yin, Neubig, and
  Martins}]{fernandes-etal-2021-measuring}
Patrick Fernandes, Kayo Yin, Graham Neubig, and Andr{\'e} F.~T. Martins. 2021.
\newblock \href {https://doi.org/10.18653/v1/2021.acl-long.505} {Measuring and
  increasing context usage in context-aware machine translation}.
\newblock In \emph{Proceedings of the 59th Annual Meeting of the Association
  for Computational Linguistics and the 11th International Joint Conference on
  Natural Language Processing (Volume 1: Long Papers)}, pages 6467--6478,
  Online. Association for Computational Linguistics.

\bibitem[{Goyal et~al.(2020)Goyal, Choudhury, Raje, Chakaravarthy, Sabharwal,
  and Verma}]{goyal2020power}
Saurabh Goyal, Anamitra~Roy Choudhury, Saurabh Raje, Venkatesan Chakaravarthy,
  Yogish Sabharwal, and Ashish Verma. 2020.
\newblock \href {https://proceedings.mlr.press/v119/goyal20a.html}
  {{P}o{WER}-{BERT}: Accelerating {BERT} inference via progressive word-vector
  elimination}.
\newblock In \emph{Proceedings of the 37th International Conference on Machine
  Learning}, volume 119 of \emph{Proceedings of Machine Learning Research},
  pages 3690--3699. PMLR.

\bibitem[{Hardmeier(2012)}]{hardmeier2012discourse}
Christian Hardmeier. 2012.
\newblock \href {https://doi.org/10.4000/discours.8726} {Discourse in
  statistical machine translation: A survey and a case study}.
\newblock \emph{Discours-Revue de linguistique, psycholinguistique et
  informatique}, 11.

\bibitem[{Hassan et~al.(2018)Hassan, Aue, Chen, Chowdhary, Clark, Federmann,
  Huang, Junczys-Dowmunt, Lewis, Li et~al.}]{hassan2018achieving}
Hany Hassan, Anthony Aue, Chang Chen, Vishal Chowdhary, Jonathan Clark,
  Christian Federmann, Xuedong Huang, Marcin Junczys-Dowmunt, William Lewis,
  Mu~Li, et~al. 2018.
\newblock Achieving human parity on automatic chinese to english news
  translation.
\newblock \emph{arXiv preprint arXiv:1803.05567}.

\bibitem[{Huo et~al.(2020)Huo, Herold, Gao, Dahlmann, Khadivi, and
  Ney}]{huo-etal-2020-diving}
Jingjing Huo, Christian Herold, Yingbo Gao, Leonard Dahlmann, Shahram Khadivi,
  and Hermann Ney. 2020.
\newblock \href {https://aclanthology.org/2020.wmt-1.71} {Diving deep into
  context-aware neural machine translation}.
\newblock In \emph{Proceedings of the Fifth Conference on Machine Translation},
  pages 604--616, Online. Association for Computational Linguistics.

\bibitem[{Hwang et~al.(2021)Hwang, Yun, and Jung}]{hwang-etal-2021-contrastive}
Yongkeun Hwang, Hyeongu Yun, and Kyomin Jung. 2021.
\newblock \href {https://aclanthology.org/2021.wmt-1.121} {Contrastive learning
  for context-aware neural machine translation using coreference information}.
\newblock In \emph{Proceedings of the Sixth Conference on Machine Translation},
  pages 1135--1144, Online. Association for Computational Linguistics.

\bibitem[{Jean et~al.(2017)Jean, Lauly, Firat, and Cho}]{jean2017does}
Sebastien Jean, Stanislas Lauly, Orhan Firat, and Kyunghyun Cho. 2017.
\newblock Does neural machine translation benefit from larger context?
\newblock \emph{arXiv preprint arXiv:1704.05135}.

\bibitem[{Kitaev et~al.(2020)Kitaev, Kaiser, and Levskaya}]{kitaev2020reformer}
Nikita Kitaev, {\L}ukasz Kaiser, and Anselm Levskaya. 2020.
\newblock Reformer: The efficient transformer.
\newblock \emph{arXiv preprint arXiv:2001.04451}.

\bibitem[{Kudo and Richardson(2018)}]{kudo-richardson-2018-sentencepiece}
Taku Kudo and John Richardson. 2018.
\newblock \href {https://doi.org/10.18653/v1/D18-2012} {{S}entence{P}iece: A
  simple and language independent subword tokenizer and detokenizer for neural
  text processing}.
\newblock In \emph{Proceedings of the 2018 Conference on Empirical Methods in
  Natural Language Processing: System Demonstrations}, pages 66--71, Brussels,
  Belgium. Association for Computational Linguistics.

\bibitem[{Li et~al.(2020)Li, Liu, Wang, Jiang, Xiao, Zhu, Liu, and
  Li}]{li-etal-2020-multi-encoder}
Bei Li, Hui Liu, Ziyang Wang, Yufan Jiang, Tong Xiao, Jingbo Zhu, Tongran Liu,
  and Changliang Li. 2020.
\newblock \href {https://doi.org/10.18653/v1/2020.acl-main.322} {Does
  multi-encoder help? a case study on context-aware neural machine
  translation}.
\newblock In \emph{Proceedings of the 58th Annual Meeting of the Association
  for Computational Linguistics}, pages 3512--3518, Online. Association for
  Computational Linguistics.

\bibitem[{Lin et~al.(2022)Lin, Wang, Liu, and Qiu}]{lin2022survey}
Tianyang Lin, Yuxin Wang, Xiangyang Liu, and Xipeng Qiu. 2022.
\newblock A survey of transformers.
\newblock \emph{AI Open}.

\bibitem[{Lison et~al.(2018)Lison, Tiedemann, and
  Kouylekov}]{lison-etal-2018-opensubtitles2018}
Pierre Lison, J{\"o}rg Tiedemann, and Milen Kouylekov. 2018.
\newblock \href {https://aclanthology.org/L18-1275} {{O}pen{S}ubtitles2018:
  Statistical rescoring of sentence alignments in large, noisy parallel
  corpora}.
\newblock In \emph{Proceedings of the Eleventh International Conference on
  Language Resources and Evaluation ({LREC} 2018)}, Miyazaki, Japan. European
  Language Resources Association (ELRA).

\bibitem[{Lopes et~al.(2020)Lopes, Farajian, Bawden, Zhang, and
  Martins}]{lopes-etal-2020-document}
Ant{\'o}nio Lopes, M.~Amin Farajian, Rachel Bawden, Michael Zhang, and
  Andr{\'e} F.~T. Martins. 2020.
\newblock \href {https://aclanthology.org/2020.eamt-1.24} {Document-level
  neural {MT}: A systematic comparison}.
\newblock In \emph{Proceedings of the 22nd Annual Conference of the European
  Association for Machine Translation}, pages 225--234, Lisboa, Portugal.
  European Association for Machine Translation.

\bibitem[{Lupo et~al.(2022)Lupo, Dinarelli, and
  Besacier}]{lupo-etal-2022-divide}
Lorenzo Lupo, Marco Dinarelli, and Laurent Besacier. 2022.
\newblock \href {https://doi.org/10.18653/v1/2022.acl-long.312} {Divide and
  rule: Effective pre-training for context-aware multi-encoder translation
  models}.
\newblock In \emph{Proceedings of the 60th Annual Meeting of the Association
  for Computational Linguistics (Volume 1: Long Papers)}, pages 4557--4572,
  Dublin, Ireland. Association for Computational Linguistics.

\bibitem[{Ma et~al.(2020)Ma, Zhang, and Zhou}]{ma-etal-2020-simple}
Shuming Ma, Dongdong Zhang, and Ming Zhou. 2020.
\newblock \href {https://doi.org/10.18653/v1/2020.acl-main.321} {A simple and
  effective unified encoder for document-level machine translation}.
\newblock In \emph{Proceedings of the 58th Annual Meeting of the Association
  for Computational Linguistics}, pages 3505--3511, Online. Association for
  Computational Linguistics.

\bibitem[{Majumde et~al.(2022)Majumde, Lauly, Nadejde, Federico, and
  Dinu}]{majumde2022baseline}
Suvodeep Majumde, Stanislas Lauly, Maria Nadejde, Marcello Federico, and
  Georgiana Dinu. 2022.
\newblock A baseline revisited: Pushing the limits of multi-segment models for
  context-aware translation.
\newblock \emph{arXiv preprint arXiv:2210.10906}.

\bibitem[{Martins and Astudillo(2016)}]{martins2016softmax}
Andr{\'e}~FT Martins and Ram{\'o}n~F Astudillo. 2016.
\newblock From softmax to sparsemax: a sparse model of attention and
  multi-label classification.
\newblock In \emph{Proceedings of the 33rd International Conference on
  International Conference on Machine Learning-Volume 48}, pages 1614--1623.

\bibitem[{Maruf et~al.(2019)Maruf, Martins, and
  Haffari}]{maruf-etal-2019-selective}
Sameen Maruf, Andr{\'e} F.~T. Martins, and Gholamreza Haffari. 2019.
\newblock \href {https://doi.org/10.18653/v1/N19-1313} {Selective attention for
  context-aware neural machine translation}.
\newblock In \emph{Proceedings of the 2019 Conference of the North {A}merican
  Chapter of the Association for Computational Linguistics: Human Language
  Technologies, Volume 1 (Long and Short Papers)}, pages 3092--3102,
  Minneapolis, Minnesota. Association for Computational Linguistics.

\bibitem[{Mathy and Feldman(2012)}]{mathy2012whats}
Fabien Mathy and Jacob Feldman. 2012.
\newblock What’s magic about magic numbers? chunking and data compression in
  short-term memory.
\newblock \emph{Cognition}, 122(3):346--362.

\bibitem[{Meister et~al.(2021)Meister, Lazov, Augenstein, and
  Cotterell}]{meister-etal-2021-sparse}
Clara Meister, Stefan Lazov, Isabelle Augenstein, and Ryan Cotterell. 2021.
\newblock \href {https://doi.org/10.18653/v1/2021.acl-short.17} {Is sparse
  attention more interpretable?}
\newblock In \emph{Proceedings of the 59th Annual Meeting of the Association
  for Computational Linguistics and the 11th International Joint Conference on
  Natural Language Processing (Volume 2: Short Papers)}, pages 122--129,
  Online. Association for Computational Linguistics.

\bibitem[{Miculicich et~al.(2018)Miculicich, Ram, Pappas, and
  Henderson}]{miculicich-etal-2018-document}
Lesly Miculicich, Dhananjay Ram, Nikolaos Pappas, and James Henderson. 2018.
\newblock \href {https://doi.org/10.18653/v1/D18-1325} {Document-level neural
  machine translation with hierarchical attention networks}.
\newblock In \emph{Proceedings of the 2018 Conference on Empirical Methods in
  Natural Language Processing}, pages 2947--2954, Brussels, Belgium.
  Association for Computational Linguistics.

\bibitem[{Miller(1956)}]{miller1956magical}
George~A Miller. 1956.
\newblock The magical number seven, plus or minus two: Some limits on our
  capacity for processing information.
\newblock \emph{Psychological review}, 63(2):81.

\bibitem[{Morishita et~al.(2021)Morishita, Suzuki, Iwata, and
  Nagata}]{morishita-etal-2021-context}
Makoto Morishita, Jun Suzuki, Tomoharu Iwata, and Masaaki Nagata. 2021.
\newblock \href {https://doi.org/10.18653/v1/2021.eacl-main.214} {Context-aware
  neural machine translation with mini-batch embedding}.
\newblock In \emph{Proceedings of the 16th Conference of the European Chapter
  of the Association for Computational Linguistics: Main Volume}, pages
  2513--2521, Online. Association for Computational Linguistics.

\bibitem[{M{\"u}ller et~al.(2018)M{\"u}ller, Rios, Voita, and
  Sennrich}]{muller-etal-2018-large}
Mathias M{\"u}ller, Annette Rios, Elena Voita, and Rico Sennrich. 2018.
\newblock \href {https://doi.org/10.18653/v1/W18-6307} {A large-scale test set
  for the evaluation of context-aware pronoun translation in neural machine
  translation}.
\newblock In \emph{Proceedings of the Third Conference on Machine Translation:
  Research Papers}, pages 61--72, Brussels, Belgium. Association for
  Computational Linguistics.

\bibitem[{Nawrot et~al.(2022)Nawrot, Tworkowski, Tyrolski, Kaiser, Wu, Szegedy,
  and Michalewski}]{nawrot-etal-2022-hierarchical}
Piotr Nawrot, Szymon Tworkowski, Micha{\l} Tyrolski, Lukasz Kaiser, Yuhuai Wu,
  Christian Szegedy, and Henryk Michalewski. 2022.
\newblock \href {https://doi.org/10.18653/v1/2022.findings-naacl.117}
  {Hierarchical transformers are more efficient language models}.
\newblock In \emph{Findings of the Association for Computational Linguistics:
  NAACL 2022}, pages 1559--1571, Seattle, United States. Association for
  Computational Linguistics.

\bibitem[{Ott et~al.(2019)Ott, Edunov, Baevski, Fan, Gross, Ng, Grangier, and
  Auli}]{ott-etal-2019-fairseq}
Myle Ott, Sergey Edunov, Alexei Baevski, Angela Fan, Sam Gross, Nathan Ng,
  David Grangier, and Michael Auli. 2019.
\newblock \href {https://doi.org/10.18653/v1/N19-4009} {fairseq: A fast,
  extensible toolkit for sequence modeling}.
\newblock In \emph{Proceedings of the 2019 Conference of the North {A}merican
  Chapter of the Association for Computational Linguistics (Demonstrations)},
  pages 48--53, Minneapolis, Minnesota. Association for Computational
  Linguistics.

\bibitem[{Papineni et~al.(2002)Papineni, Roukos, Ward, and
  Zhu}]{papineni-etal-2002-bleu}
Kishore Papineni, Salim Roukos, Todd Ward, and Wei-Jing Zhu. 2002.
\newblock \href {https://doi.org/10.3115/1073083.1073135} {{B}leu: a method for
  automatic evaluation of machine translation}.
\newblock In \emph{Proceedings of the 40th Annual Meeting of the Association
  for Computational Linguistics}, pages 311--318, Philadelphia, Pennsylvania,
  USA. Association for Computational Linguistics.

\bibitem[{Post(2018)}]{post-2018-call}
Matt Post. 2018.
\newblock \href {https://doi.org/10.18653/v1/W18-6319} {A call for clarity in
  reporting {BLEU} scores}.
\newblock In \emph{Proceedings of the Third Conference on Machine Translation:
  Research Papers}, pages 186--191, Brussels, Belgium. Association for
  Computational Linguistics.

\bibitem[{Rei et~al.(2022)Rei, C.~de Souza, Alves, Zerva, Farinha, Glushkova,
  Lavie, Coheur, and Martins}]{rei-etal-2022-comet}
Ricardo Rei, Jos{\'e}~G. C.~de Souza, Duarte Alves, Chrysoula Zerva, Ana~C
  Farinha, Taisiya Glushkova, Alon Lavie, Luisa Coheur, and Andr{\'e} F.~T.
  Martins. 2022.
\newblock \href {https://aclanthology.org/2022.wmt-1.52} {{COMET}-22:
  Unbabel-{IST} 2022 submission for the metrics shared task}.
\newblock In \emph{Proceedings of the Seventh Conference on Machine Translation
  (WMT)}, pages 578--585, Abu Dhabi, United Arab Emirates (Hybrid). Association
  for Computational Linguistics.

\bibitem[{Rei et~al.(2020)Rei, Stewart, Farinha, and
  Lavie}]{rei-etal-2020-comet}
Ricardo Rei, Craig Stewart, Ana~C Farinha, and Alon Lavie. 2020.
\newblock \href {https://doi.org/10.18653/v1/2020.emnlp-main.213} {{COMET}: A
  neural framework for {MT} evaluation}.
\newblock In \emph{Proceedings of the 2020 Conference on Empirical Methods in
  Natural Language Processing (EMNLP)}, pages 2685--2702, Online. Association
  for Computational Linguistics.

\bibitem[{Sennrich et~al.(2016)Sennrich, Haddow, and
  Birch}]{sennrich-etal-2016-neural}
Rico Sennrich, Barry Haddow, and Alexandra Birch. 2016.
\newblock \href {https://doi.org/10.18653/v1/P16-1162} {Neural machine
  translation of rare words with subword units}.
\newblock In \emph{Proceedings of the 54th Annual Meeting of the Association
  for Computational Linguistics (Volume 1: Long Papers)}, pages 1715--1725,
  Berlin, Germany. Association for Computational Linguistics.

\bibitem[{Subramanian et~al.(2020)Subramanian, Collobert, Ranzato, and
  Boureau}]{subramanian2020multi}
Sandeep Subramanian, Ronan Collobert, Marc'Aurelio Ranzato, and Y-Lan Boureau.
  2020.
\newblock Multi-scale transformer language models.
\newblock \emph{arXiv preprint arXiv:2005.00581}.

\bibitem[{Sun et~al.(2022)Sun, Wang, Zhou, Zhao, Huang, Chen, and
  Li}]{sun-etal-2022-rethinking}
Zewei Sun, Mingxuan Wang, Hao Zhou, Chengqi Zhao, Shujian Huang, Jiajun Chen,
  and Lei Li. 2022.
\newblock \href {https://doi.org/10.18653/v1/2022.findings-acl.279} {Rethinking
  document-level neural machine translation}.
\newblock In \emph{Findings of the Association for Computational Linguistics:
  ACL 2022}, pages 3537--3548, Dublin, Ireland. Association for Computational
  Linguistics.

\bibitem[{Tay et~al.(2022)Tay, Dehghani, Bahri, and Metzler}]{tay2022efficient}
Yi~Tay, Mostafa Dehghani, Dara Bahri, and Donald Metzler. 2022.
\newblock \href {https://doi.org/10.1145/3530811} {Efficient transformers: A
  survey}.
\newblock \emph{ACM Comput. Surv.}, 55(6).

\bibitem[{Tay et~al.(2021)Tay, Tran, Ruder, Gupta, Chung, Bahri, Qin,
  Baumgartner, Yu, and Metzler}]{tay2021charformer}
Yi~Tay, Vinh~Q Tran, Sebastian Ruder, Jai Gupta, Hyung~Won Chung, Dara Bahri,
  Zhen Qin, Simon Baumgartner, Cong Yu, and Donald Metzler. 2021.
\newblock Charformer: Fast character transformers via gradient-based subword
  tokenization.
\newblock \emph{arXiv preprint arXiv:2106.12672}.

\bibitem[{Terrace(2002)}]{terrace2002comparative}
H.~S. Terrace. 2002.
\newblock \href {https://doi.org/10.1007/978-1-4615-0821-2_2} {\emph{The
  Comparative Psychology of Chunking}}, pages 23--55. Springer US, Boston, MA.

\bibitem[{Tiedemann et~al.(2022)Tiedemann, Aulamo, Bakshandaeva, Boggia,
  Gr{\"o}nroos, Nieminen, Raganato, Scherrer, Vazquez, and
  Virpioja}]{tiedemann2022democratizing}
J{\"o}rg Tiedemann, Mikko Aulamo, Daria Bakshandaeva, Michele Boggia, Stig-Arne
  Gr{\"o}nroos, Tommi Nieminen, Alessandro Raganato, Yves Scherrer, Raul
  Vazquez, and Sami Virpioja. 2022.
\newblock Democratizing machine translation with opus-mt.
\newblock \emph{arXiv preprint arXiv:2212.01936}.

\bibitem[{Tiedemann and Scherrer(2017)}]{tiedemann-scherrer-2017-neural}
J{\"o}rg Tiedemann and Yves Scherrer. 2017.
\newblock \href {https://doi.org/10.18653/v1/W17-4811} {Neural machine
  translation with extended context}.
\newblock In \emph{Proceedings of the Third Workshop on Discourse in Machine
  Translation}, pages 82--92, Copenhagen, Denmark. Association for
  Computational Linguistics.

\bibitem[{Tu et~al.(2017)Tu, Liu, Lu, Liu, and Li}]{tu-etal-2017-context}
Zhaopeng Tu, Yang Liu, Zhengdong Lu, Xiaohua Liu, and Hang Li. 2017.
\newblock \href {https://doi.org/10.1162/tacl_a_00048} {Context gates for
  neural machine translation}.
\newblock \emph{Transactions of the Association for Computational Linguistics},
  5:87--99.

\bibitem[{Tu et~al.(2018)Tu, Liu, Shi, and Zhang}]{tu-etal-2018-learning}
Zhaopeng Tu, Yang Liu, Shuming Shi, and Tong Zhang. 2018.
\newblock \href {https://doi.org/10.1162/tacl_a_00029} {Learning to remember
  translation history with a continuous cache}.
\newblock \emph{Transactions of the Association for Computational Linguistics},
  6:407--420.

\bibitem[{Vaswani et~al.(2017)Vaswani, Shazeer, Parmar, Uszkoreit, Jones,
  Gomez, Kaiser, and Polosukhin}]{vaswani2017attention}
Ashish Vaswani, Noam Shazeer, Niki Parmar, Jakob Uszkoreit, Llion Jones,
  Aidan~N Gomez, {\L}ukasz Kaiser, and Illia Polosukhin. 2017.
\newblock Attention is all you need.
\newblock \emph{Advances in neural information processing systems}, 30.

\bibitem[{Voita et~al.(2019{\natexlab{a}})Voita, Sennrich, and
  Titov}]{voita-etal-2019-context}
Elena Voita, Rico Sennrich, and Ivan Titov. 2019{\natexlab{a}}.
\newblock \href {https://doi.org/10.18653/v1/D19-1081} {Context-aware
  monolingual repair for neural machine translation}.
\newblock In \emph{Proceedings of the 2019 Conference on Empirical Methods in
  Natural Language Processing and the 9th International Joint Conference on
  Natural Language Processing (EMNLP-IJCNLP)}, pages 877--886, Hong Kong,
  China. Association for Computational Linguistics.

\bibitem[{Voita et~al.(2019{\natexlab{b}})Voita, Sennrich, and
  Titov}]{voita-etal-2019-good}
Elena Voita, Rico Sennrich, and Ivan Titov. 2019{\natexlab{b}}.
\newblock \href {https://doi.org/10.18653/v1/P19-1116} {When a good translation
  is wrong in context: Context-aware machine translation improves on deixis,
  ellipsis, and lexical cohesion}.
\newblock In \emph{Proceedings of the 57th Annual Meeting of the Association
  for Computational Linguistics}, pages 1198--1212, Florence, Italy.
  Association for Computational Linguistics.

\bibitem[{Voita et~al.(2018)Voita, Serdyukov, Sennrich, and
  Titov}]{voita-etal-2018-context}
Elena Voita, Pavel Serdyukov, Rico Sennrich, and Ivan Titov. 2018.
\newblock \href {https://doi.org/10.18653/v1/P18-1117} {Context-aware neural
  machine translation learns anaphora resolution}.
\newblock In \emph{Proceedings of the 56th Annual Meeting of the Association
  for Computational Linguistics (Volume 1: Long Papers)}, pages 1264--1274,
  Melbourne, Australia. Association for Computational Linguistics.

\bibitem[{Wang et~al.(2020)Wang, Li, Khabsa, Fang, and Ma}]{wang2020linformer}
Sinong Wang, Belinda~Z Li, Madian Khabsa, Han Fang, and Hao Ma. 2020.
\newblock Linformer: Self-attention with linear complexity.
\newblock \emph{arXiv preprint arXiv:2006.04768}.

\bibitem[{Wong and Kit(2012)}]{wong-kit-2012-extending}
Billy T.~M. Wong and Chunyu Kit. 2012.
\newblock \href {https://aclanthology.org/D12-1097} {Extending machine
  translation evaluation metrics with lexical cohesion to document level}.
\newblock In \emph{Proceedings of the 2012 Joint Conference on Empirical
  Methods in Natural Language Processing and Computational Natural Language
  Learning}, pages 1060--1068, Jeju Island, Korea. Association for
  Computational Linguistics.

\bibitem[{Wu et~al.(2022)Wu, Xia, Zhu, Wu, Xie, and Qin}]{wu2022study}
Xueqing Wu, Yingce Xia, Jinhua Zhu, Lijun Wu, Shufang Xie, and Tao Qin. 2022.
\newblock \href {https://doi.org/10.1007/s10994-021-06070-y} {A study of bert
  for context-aware neural machine translation}.
\newblock \emph{Machine Learning}, 111(3):917--935.

\bibitem[{Yin et~al.(2021)Yin, Fernandes, Pruthi, Chaudhary, Martins, and
  Neubig}]{yin-etal-2021-context}
Kayo Yin, Patrick Fernandes, Danish Pruthi, Aditi Chaudhary, Andr{\'e} F.~T.
  Martins, and Graham Neubig. 2021.
\newblock \href {https://doi.org/10.18653/v1/2021.acl-long.65} {Do
  context-aware translation models pay the right attention?}
\newblock In \emph{Proceedings of the 59th Annual Meeting of the Association
  for Computational Linguistics and the 11th International Joint Conference on
  Natural Language Processing (Volume 1: Long Papers)}, pages 788--801, Online.
  Association for Computational Linguistics.

\bibitem[{Zhang et~al.(2020)Zhang, Chen, Ge, and Fan}]{zhang-etal-2020-long}
Pei Zhang, Boxing Chen, Niyu Ge, and Kai Fan. 2020.
\newblock \href {https://doi.org/10.18653/v1/2020.emnlp-main.81} {Long-short
  term masking transformer: A simple but effective baseline for document-level
  neural machine translation}.
\newblock In \emph{Proceedings of the 2020 Conference on Empirical Methods in
  Natural Language Processing (EMNLP)}, pages 1081--1087, Online. Association
  for Computational Linguistics.

\bibitem[{Zheng et~al.(2021)Zheng, Yue, Huang, Chen, and
  Birch}]{zheng2021towards}
Zaixiang Zheng, Xiang Yue, Shujian Huang, Jiajun Chen, and Alexandra Birch.
  2021.
\newblock Towards making the most of context in neural machine translation.
\newblock In \emph{Proceedings of the Twenty-Ninth International Conference on
  International Joint Conferences on Artificial Intelligence}, pages
  3983--3989.

\end{thebibliography}
\bibliographystyle{acl_natbib}

\appendix

\section{Models and Training Details}
\label{sec:models-and-training-details}

To implement and train our models we used fairseq framework \citep{ott-etal-2019-fairseq} and based our code on the codebase of \citet{fernandes-etal-2021-measuring}. All models were based on the transformer-base configuration. The shared hyper-parameters are presented in Table~\ref{tab:hyper-parameters}. We trained each model on a single GPU (NVIDIA GeForce RTX 3090 24GB).

For Latent Grouping and Shortening, we used a categorizing FFN with 512 hidden units, the number of inputs equal to the Embed Dim, and the number of outputs equal to the number of groups. 
Table~\ref{tab:models-details} shows the number of parameters for each model.

\begin{table}[!ht]
    \centering
    \begin{tabular}{lr}
    \hline
        \textbf{Hyper-parameter} & \textbf{Value}    \\ \hline
        Encoder Layers & 6 \\ 
        Decoder Layers  & 6 \\ 
        Attention Heads  & 8 \\ 
        Embed Dim  & 512 \\ 
        FFN Embed Dim & 2048 \\
        Dropout & 0.3 \\ 
        Share Decoder In/Out Embed & True \\ 
        Optimizer & Adam \\
        Adam Betas & 0.9, 0.98 \\
        Adam Epsilon & 1e-8 \\
        Learning Rate & 5e-4 \\
        LR Scheduler & Inverse Sqrt \\
        LR Warmup Updates & 2500 \\
        Weight Decay & 0.0001 \\
        Label Smoothing & 0.1 \\
        Clip Norm & 0.1 \\
        Batch Max Tokens & 4096 \\
        Update Frequency & 8 \\
        Max Epoch & - \\
        Patience & 5 \\
        Beam & 5 \\
        Max Vocab Size & 20000 \\
        Seed & 42 \\
        \hline
    \end{tabular}
    \caption{The shared hyper-parameters of the tested models.}
    \label{tab:hyper-parameters}
\end{table}

\begin{table}[!ht]
    \centering
    \begin{tabular}{lr}
    \hline
        \textbf{Model} & \textbf{Parameters}    \\ \hline
        Sentence-level & 64.42M   \\ 
        Single-encoder  & 64.42M     \\ 
        Multi-encoder &  83.33M   \\ 
        Caching Tokens & 71.25M   \\ 
        Caching Sentence & 71.26M  \\ 
        Shortening - Max Pooling & 72.83M  \\ 
        Shortening - Avg Pooling & 72.83M \\ 
        Shortening - Linear Pooling & 73.35M \\ 
        Shortening - Grouping & 72.58M \\ 
        Shortening - Selecting & 72.58M  \\
        \hline
    \end{tabular}
    \caption{The number of parameters in the tested models.}
    \label{tab:models-details}
\end{table}

\section{COMET Results}
\label{sec:additional-result}

Apart from BLEU and contrastive dataset accuracy presented in Section~\ref{sec:experiments}, we also measured COMET \citep{rei-etal-2020-comet} based on \texttt{Unbabel/wmt22-comet-da} model \citep{rei-etal-2022-comet}. See Tables~\ref{tab:results-additional-iwslt-en-de} and \ref{tab:results-additional-iwslt-en-fr} for the results on English-to-German and English-to-French respectively.

\begin{table*}[!ht]
    \centering
    \begin{tabular}{lrrrrrr}
    \hline
        \textbf{Model} & \textbf{Context: 0} \textbf{} & \textbf{}    \\ \hline
        Sentence-level & 0.7778 & ~ & ~   \\ 
        \hline
        \textbf{Model} & \textbf{Context: 1} & \textbf{Context: 2} & \textbf{Context: 3}   \\ 
        \hline
        Single-encoder & 0.7831 & 0.7789 & 0.7758    \\ 
        Multi-encoder & 0.7831 & \textbf{0.7871} & \textbf{0.7856}   \\ 
        \hline
        Caching Tokens & 0.7806 & 0.7776 & 0.7821  \\ 
        Caching Sentence & 0.7712 & 0.7640 & 0.7673   \\ 
        Shortening - Max Pooling & 0.7743 & 0.7772 & 0.7799   \\ 
        Shortening - Avg Pooling & 0.7774 & 0.7770 & 0.7844  \\ 
        Shortening - Linear Pooling & 0.7757 & 0.7745 & 0.7823   \\ 
        Shortening - Grouping & \textbf{0.7842} & 0.7828 & 0.7811  \\ 
        Shortening - Selecting & 0.7774 & 0.7826 & 0.7836  \\
        \hline
    \end{tabular}
    \caption{Results in terms of COMET \citep{rei-etal-2020-comet} based on \texttt{Unbabel/wmt22-comet-da} model \citep{rei-etal-2022-comet} of the \textbf{En-De} IWSLT 2017 experiment.}
    \label{tab:results-additional-iwslt-en-de}
\end{table*}

\begin{table*}[!ht]
    \centering
    \begin{tabular}{lrrrrrr}
    \hline
        \textbf{Model} & \textbf{Context: 0} \textbf{} & \textbf{}    \\ \hline
        Sentence-level & 0.7943 & ~ & ~   \\ 
        \hline
        \textbf{Model} & \textbf{Context: 1} & \textbf{Context: 2} & \textbf{Context: 3}   \\ 
        \hline
        Single-encoder  & 0.7930 & \textbf{0.7979} & 0.7913    \\ 
        Multi-encoder & \textbf{0.7968} & 0.7934 & 0.7934   \\ 
        \hline
        Caching Tokens & 0.7923 & 0.7935 & 0.7945  \\ 
        Caching Sentence & 0.7845 & 0.7654 & 0.7737   \\ 
        Shortening - Max Pooling & 0.7911 & 0.7913 & \textbf{0.7974}   \\ 
        Shortening - Avg Pooling & 0.7920 & 0.7924 & 0.7952  \\ 
        Shortening - Linear Pooling & 0.7933 & 0.7951 & 0.7927   \\ 
        Shortening - Grouping & 0.7933 & 0.7976 & 0.7921  \\ 
        Shortening - Selecting & 0.7951 & 0.7945 & 0.7935  \\
        \hline
    \end{tabular}
    \caption{Results in terms of  COMET \citep{rei-etal-2020-comet} based on \texttt{Unbabel/wmt22-comet-da} model \citep{rei-etal-2022-comet} of the \textbf{En-Fr} IWSLT 2017 experiment.}
    \label{tab:results-additional-iwslt-en-fr}
\end{table*}

\section{Detailed Contrastive Results}
\label{sec:detailed-contrastive-result}

In this section we report the accuracy on the contrastive datasets for the different placements of the antecedent. The antecedent distance of zero corresponds to the examples where the antecedent is in the current sentence. The value of one represent the antecedent in the first context sentence (counting backward from the current sentence), etc. The results of the ContraPro dataset (English-to-German) and the contrastive dataset by \citet{lopes-etal-2020-document} (English-to-French) are presented in Tables~\ref{tab:results-contrapro-detailed-iwslt-en-de} and \ref{tab:results-contrapro-detailed-iwslt-en-fr} respectively.

\begin{table*}
    \centering
    \begin{tabular}{lcccccc}
        \hline
         &  &  \multicolumn{5}{c}{\textbf{Antecedent Distance}} \\
         \textbf{Model} &  \textbf{Context} &  \textbf{0} &  \textbf{1} &  \textbf{2} &  \textbf{3} & \textbf{>3} \\
         \hline
         Sentence-level&  0&  72.21\%&  31.82\%&  44.90\%&  48.87\%& 67.42\%\\
         \hline
         Single-encoder&  1&  70.08\%&  38.42\%&  46.16\%&  49.04\%& 70.59\%\\
         &  2&  73.96\%&  37.87\%&  48.48\%&  50.79\%& 69.00\%\\
         &  3&  71.79\%&  40.00\%&  47.88\%&  52.01\%& 66.06\%\\
         \hline
         Multi-encoder&  1&  75.17\%&  33.16\%&  44.64\%&  47.47\%& 66.97\%\\
         &  2&  73.54\%&  35.63\%&  47.42\%&  50.79\%& 69.00\%\\
         &  3&  70.88\%&  33.99\%&  46.16\%&  50.61\%& 69.46\%\\
         \hline
         Caching Tokens&  1&  72.21\%&  49.07\%&  45.03\%&  50.09\%& 71.27\%\\
         & 2& 70.75\%& 47.17\%& 58.74\%& 48.69\%&66.74\%\\
         & 3& 70.25\%& 42.53\%& 52.98\%& 60.91\%&68.78\%\\
         \hline
         Caching Sentence& 1& 66.83\%& 36.78\%& 45.63\%& 50.26\%&68.55\%\\
         & 2& 66.83\%& 35.42\%& 47.81\%& 49.74\%&71.04\%\\
         & 3& 60.17\%& 37.16\%& 47.95\%& 50.96\%&67.87\%\\
         \hline
         Shortening - Max Pooling& 1& 68.92\%& 46.33\%& 44.64\%& 48.17\%&71.95\%\\
         & 2& 72.83\%& 47.63\%& 62.12\%& 47.64\%&63.57\%\\
         & 3& 72.13\%& 40.83\%& 53.71\%& 63.00\%&71.27\%\\
         \hline
         Shortening - Avg Pooling& 1& 70.04\%& 48.58\%& 45.50\%& 48.52\%&72.62\%\\
         & 2& 72.67\%& 47.04\%& 62.58\%& 47.64\%&64.93\%\\
         & 3& 70.88\%& 40.71\%& 54.24\%& 60.56\%&71.95\%\\
         \hline
         Shortening - Linear Pooling& 1& 69.13\%& 47.84\%& 44.64\%& 49.21\%&73.53\%\\
         & 2& 70.38\%& 43.75\%& 59.34\%& 47.99\%&67.87\%\\
         & 3& 72.58\%& 41.06\%& 54.90\%& 64.05\%&69.91\%\\
         \hline
         Shortening - Grouping& 1& 73.67\%& 53.64\%& 45.56\%& 46.95\%&71.72\%\\
         & 2& 69.17\%& 47.66\%& 61.85\%& 47.29\%&68.78\%\\
         & 3& 71.21\%& 41.58\%& 55.03\%& 62.13\%&68.10\%\\
         \hline
         Shortening - Selecting& 1& 72.88\%& 50.16\%& 43.77\%& 47.64\%&69.00\%\\
         & 2& 71.75\%& 45.85\%& 64.04\%& 47.99\%&67.19\%\\
         & 3& 73.29\%& 42.04\%& 54.57\%& 65.10\%&68.78\%\\
         \hline
    \end{tabular}
    \caption{Detailed results of the accuracy on the ContraPro contrastive dataset for different antecedent locations of the models trained on the \textbf{En-De} IWSLT 2017 dataset.}
    \label{tab:results-contrapro-detailed-iwslt-en-de}
\end{table*}

\begin{table*}
    \centering
    \begin{tabular}{lcccccc}
        \hline
         &  &  \multicolumn{5}{c}{\textbf{Antecedent Distance}} \\
         \textbf{Model} &  \textbf{Context} &  \textbf{0} &  \textbf{1} &  \textbf{2} &  \textbf{3} & \textbf{>3} \\
         \hline
         Sentence-level&  0&  75.86\%&  75.76\%&  76.98\%&  76.70\%& 74.55\%\\
         \hline
         Single-encoder&  1&  76.88\%&  77.16\%&  78.39\%&  78.86\%& 76.89\%\\
         &  2&  78.92\%&  78.69\%&  80.17\%&  78.98\%& 78.70\%\\
         &  3&  80.37\%&  80.99\%&  81.77\%&  81.70\%& 81.15\%\\
         \hline
         Multi-encoder&  1&  75.71\%&  75.08\%&  76.92\%&  76.93\%& 75.72\%\\
         &  2&  77.03\%&  77.16\%&  78.08\%&  78.52\%& 76.14\%\\
         &  3&  75.08\%&  76.11\%&  77.53\%&  77.27\%& 74.01\%\\
         \hline
         Caching Tokens&  1&  79.80\%&  79.11\%&  80.72\%&  80.68\%& 78.81\%\\
         & 2& 79.77\%& 80.44\%& 80.79\%& 81.25\%&78.81\%\\
         & 3& 79.27\%& 80.27\%& 81.28\%& 80.34\%&79.34\%\\
         \hline
         Caching Sentence& 1& 76.81\%& 77.18\%& 78.21\%& 80.23\%&77.10\%\\
         & 2& 75.73\%& 76.52\%& 76.98\%& 78.64\%&74.76\%\\
         & 3& 75.01\%& 75.87\%& 78.27\%& 75.68\%&74.97\%\\
         \hline
         Shortening - Max Pooling& 1& 80.49\%& 80.38\%& 80.11\%& 80.11\%&81.90\%\\
         & 2& 80.25\%& 80.73\%& 81.15\%& 80.68\%&81.04\%\\
         & 3& 78.98\%& 80.27\%& 81.28\%& 80.80\%&77.96\%\\
         \hline
         Shortening - Avg Pooling& 1& 77.36\%& 77.70\%& 79.19\%& 77.61\%&78.06\%\\
         & 2& 79.94\%& 80.00\%& 80.79\%& 81.70\%&79.77\%\\
         & 3& 79.94\%& 80.77\%& 81.65\%& 81.02\%&79.02\%\\
         \hline
         Shortening - Linear Pooling& 1& 79.87\%& 80.35\%& 82.44\%& 80.57\%&81.36\%\\
         & 2& 78.80\%& 79.06\%& 80.72\%& 80.68\%&80.94\%\\
         & 3& 79.03\%& 80.09\%& 80.85\%& 80.23\%&78.59\%\\
         \hline
         Shortening - Grouping& 1& 79.28\%& 80.40\%& 81.46\%& 79.89\%&78.91\%\\
         & 2& 80.91\%& 81.30\%& 81.65\%& 81.36\%&80.62\%\\
         & 3& 79.07\%& 80.20\%& 78.88\%& 81.14\%&78.91\%\\
         \hline
         Shortening - Selecting& 1& 80.30\%& 81.03\%& 81.89\%& 82.73\%&80.40\%\\
         & 2& 80.17\%& 80.33\%& 81.40\%& 81.02\%&78.70\%\\
         & 3& 79.28\%& 80.33\%& 81.89\%& 79.55\%&81.36\%\\
         \hline
    \end{tabular}
    \caption{Detailed results of the accuracy on the contrastive dataset by \citet{lopes-etal-2020-document} for different antecedent locations of the models trained on the \textbf{En-Fr} IWSLT 2017 dataset.}
    \label{tab:results-contrapro-detailed-iwslt-en-fr}
\end{table*}

\section{Examples of Translations}
\label{sec:examples-results}

We present the examples of the translation of the sentence-level Transformer, and Selecting and Grouping Shortening architectures on the IWSLT 2017 English-to-German dataset in Table~\ref{tab:example-translations}. We marked the pronoun disambiguation from context sentences.

\begin{table*}
    \centering
    \begin{tabular}{|l|p{0.72\linewidth}|}
    \hline
         Source Context & This is a nice \ante{building}.\\
         Source Sentence & But \pronoun{it} doesn't have much to do with what a library actually does today.\\
         \hline
         Target Reference & Aber \correct{es} hat nicht viel mit dem zu tun, was eine Bibliothek heute leistet.\\
         \hline
         Sentence-level & Aber \correct{es} hat nicht viel damit zu tun, was eine Bibliothek heute tut.\\
    \hline
         Shortening - Selecting & Aber \correct{es} hat nicht viel mit der heutigen Bibliothek zu tun.\\
    \hline
         Shortening - Grouping & Aber \correct{es} hat nicht viel mit dem zu tun, was eine Bibliothek heute tut.\\
     \hline
     \hline
         Source Context & Zak Ebrahim is not my real \ante{name}.\\
         Source Sentence & I changed \pronoun{it} when my family decided to end our connection with my father and start a new life.\\
         \hline
         Target Reference & Ich habe \pronoun{ihn} geändert, als meine Familie beschloss, den Kontakt zu meinem Vater abzubrechen und ein neues Leben zu beginnen.\\
         \hline
         Sentence-level & Ich änderte \incorect{es}, als meine Familie entschied, unsere Verbindung mit meinem Vater zu beenden und ein neues Leben zu starten.\\
    \hline
         Shortening - Selecting & Ich habe \correct{ihn} verändert, als meine Familie entschied, unsere Verbindung mit meinem Vater zu beenden und ein neues Leben zu beginnen.\\
    \hline
         Shortening - Grouping & Ich habe \incorect{es} verändert, als meine Familie beschloss, unsere Verbindung mit meinem Vater zu beenden und ein neues Leben zu beginnen.\\
    \hline
    \hline
         Source Context & And this \ante{work} has been wonderful. It's been great.\\
         Source Sentence & But \pronoun{it} also has some fundamental limitations so far.\\
         \hline
         Target Reference & Aber \pronoun{sie} hat auch noch immer einige grundlegende Grenzen.\\
         \hline
         Sentence-level & Aber \incorect{es} hat bis jetzt auch einige fundamentale Grenzen.\\
    \hline
         Shortening - Selecting & Aber \incorect{es} hat bis jetzt noch grundlegende Grenzen.\\
    \hline
         Shortening - Grouping & Aber \correct{sie} hat auch bis jetzt einige fundamentale Grenzen.\\
    \hline
    \end{tabular}
    \caption{Example translations of sentence-level Transformer and Grouping and Selecting shortening context-aware models of the English sentence with the context size of one to German. We marked \ante{antecedent} and \pronoun{pronoun} in the source sentence and \correct{correct} and \incorect{incorrect} pronoun translations.}
    \label{tab:example-translations}
\end{table*}

\section{Larger Context Results}
\label{sec:larger-context-result}

In order to examine the behavior of the tested models in response to larger contexts, we trained the models on the IWSLT 2017 English-to-German dataset with context sizes up to 10. We present the results in terms of BLEU, accuracy on the ContraPro contrastive dataset, and COMET in Tables~\ref{tab:results-large-context-bleu-iwslt-en-de}, \ref{tab:results-large-context-contrapro-iwslt-en-de}, and \ref{tab:results-large-context-comet-iwslt-en-de} respectively.

\begin{table*}[!ht]
    \centering
    \begin{tabular}{lrrrrrrr}
    \hline
         & \multicolumn{7}{c}{\textbf{Context Size}} \\ 
        \textbf{Model} & \multicolumn{1}{c}{\textbf{4}} & \multicolumn{1}{c}{\textbf{5}} & \multicolumn{1}{c}{\textbf{6}} & \multicolumn{1}{c}{\textbf{7}} & \multicolumn{1}{c}{\textbf{8}} & \multicolumn{1}{c}{\textbf{9}} & \multicolumn{1}{c}{\textbf{10}}   \\ 
        \hline
        Single-encoder & 10.60 & 24.89 & 1.99 & 1.64 & 1.43 & 1.18 & 0.95    \\ 
        Multi-encoder & 28.49 & 28.34 & 27.58 & 26.69 & 25.23 & 8.76 & 7.10   \\ 
        \hline
        Caching Tokens & 28.75 & \textbf{28.61} & 27.67 & 27.90 & 27.22 & 27.15 & 26.24  \\ 
        Caching Sentence & 27.87 & 28.30 & 27.55 & 27.67 & 27.20 & 25.87 & 5.84   \\ 
        Shortening - Max Pooling & 28.32 & 28.42 & 28.15 & 28.06 & 28.03 & 28.25 & \textbf{28.53}   \\ 
        Shortening - Avg Pooling & 28.33 & 27.66 & \textbf{28.68} & 28.21 & 28.29 & \textbf{28.35} & 28.52  \\ 
        Shortening - Linear Pooling & 28.83 & 27.91 & 28.17 & \textbf{28.44} & \textbf{28.24} & 28.28 & 28.05   \\ 
        Shortening - Grouping & 28.73 & 28.15 & 28.27 & 28.21 & 27.85 & 27.65 & 28.10  \\ 
        Shortening - Selecting & \textbf{28.85} & 28.15 & 27.93 & 28.18 & 27.67 & 28.04 & 28.23  \\
        \hline
    \end{tabular}
    \caption{Results in terms of BLEU of the \textbf{En-De} IWSLT 2017 experiment for larger context sizes.}
    \label{tab:results-large-context-bleu-iwslt-en-de}
\end{table*}

\begin{table*}[!ht]
    \centering
    \begin{tabular}{lrrrrrrr}
    \hline
         & \multicolumn{7}{c}{\textbf{Context Size}} \\ 
        \textbf{Model} & \multicolumn{1}{c}{\textbf{4}} & \multicolumn{1}{c}{\textbf{5}} & \multicolumn{1}{c}{\textbf{6}} & \multicolumn{1}{c}{\textbf{7}} & \multicolumn{1}{c}{\textbf{8}} & \multicolumn{1}{c}{\textbf{9}} & \multicolumn{1}{c}{\textbf{10}}   \\ 
        \hline
        Single-encoder & 46.09\% & 44.03\% & 43.05\% & 42.07\% & 42.00\% & 38.49\% & 37.03\%   \\ 
        Multi-encoder & 47.02\% & 44.92\% & 46.25\% & 46.48\% & 43.63\% & 41.53\% & 41.44\%   \\ 
        \hline
        Caching Tokens & \textbf{53.54\%} & 47.68\% & 46.88\% & 47.04\% & 45.79\% & \textbf{48.15\%} & \textbf{48.88\%}  \\ 
        Caching Sentence & 46.57\% & 46.20\% & 44.59\% & 44.91\% & 43.29\% & 41.03\% & 43.01\%   \\ 
        Shortening - Max P. & 51.75\% & 47.13\% & 46.78\% & 46.73\% & 46.38\% & 46.38\% & 45.03\%   \\ 
        Shortening - Avg P. & 49.53\% & \textbf{49.43\%} & 47.90\% & 45.88\% & 45.59\% & 46.27\% & 44.66\%  \\ 
        Shortening - Linear P. & 48.45\% & 46.40\% & \textbf{49.31\%} & 46.35\% & 46.90\% & 45.23\% & 45.79\%   \\ 
        Shortening - Grouping & 49.55\% & 46.06\% & 45.10\% & \textbf{47.66\%} & \textbf{47.19\%} & 46.47\% & 46.53\%  \\ 
        Shortening - Selecting & 47.88\% & 48.98\% & 47.58\% & 45.58\% & 45.91\% & 45.52\% & 47.43\%  \\
        \hline
    \end{tabular}
    \caption{Results in terms of the accuracy on the ContraPro contrastive dataset of the models trained on the \textbf{En-De} IWSLT 2017 dataset for larger context sizes.}
    \label{tab:results-large-context-contrapro-iwslt-en-de}
\end{table*}

\begin{table*}[!ht]
    \centering
    \begin{tabular}{lrrrrrrr}
    \hline
         & \multicolumn{7}{c}{\textbf{Context Size}} \\ 
        \textbf{Model} & \multicolumn{1}{c}{\textbf{4}} & \multicolumn{1}{c}{\textbf{5}} & \multicolumn{1}{c}{\textbf{6}} & \multicolumn{1}{c}{\textbf{7}} & \multicolumn{1}{c}{\textbf{8}} & \multicolumn{1}{c}{\textbf{9}} & \multicolumn{1}{c}{\textbf{10}}   \\ 
        \hline
        Single-encoder & 0.6266 & 0.7376 & 0.4425 & 0.4253 & 0.3950 & 0.3738 & 0.3597    \\ 
        Multi-encoder & \textbf{0.7830} & 0.7809 & 0.7692 & 0.7621 & 0.7280 & 0.5682 & 0.5187   \\ 
        \hline
        Caching Tokens & 0.7824 & \textbf{0.7826} & 0.7773 & 0.7744 & 0.7682 & 0.7560 & 0.7450  \\ 
        Caching Sentence & 0.7766 & 0.7741 & 0.7680 & 0.7680 & 0.7637 & 0.7413 & 0.5403   \\ 
        Shortening - Max Pooling & 0.7784 & 0.7782 & 0.7799 & 0.7804 & \textbf{0.7824} & \textbf{0.7825} & 0.7790   \\ 
        Shortening - Avg Pooling & 0.7815 & 0.7806 & \textbf{0.7812} & 0.7812 & 0.7776 & 0.7781 & \textbf{0.7814}  \\ 
        Shortening - Linear Pooling & 0.7803 & 0.7810 & 0.7802 & \textbf{0.7816} & 0.7780 & 0.7808 & 0.7783   \\ 
        Shortening - Grouping & 0.7815 & 0.7808 & 0.7794 & 0.7742 & 0.7785 & 0.7757 & 0.7789  \\ 
        Shortening - Selecting & 0.7811 & 0.7793 & 0.7782 & 0.7771 & 0.7759 & 0.7750 & 0.7791  \\
        \hline
    \end{tabular}
    \caption{Results in terms of COMET \citep{rei-etal-2020-comet} based on \texttt{Unbabel/wmt22-comet-da} model \citep{rei-etal-2022-comet} of the \textbf{En-De} IWSLT 2017 experiment for larger context sizes.}
    \label{tab:results-large-context-comet-iwslt-en-de}
\end{table*}

\section{Groupings and Selections Visualization}
\label{sec:exteded-visualizaitons}

\begin{figure*}[!ht]
\center{}
    \begin{subfigure}{0.4\linewidth}
        \includegraphics[width=1\linewidth]{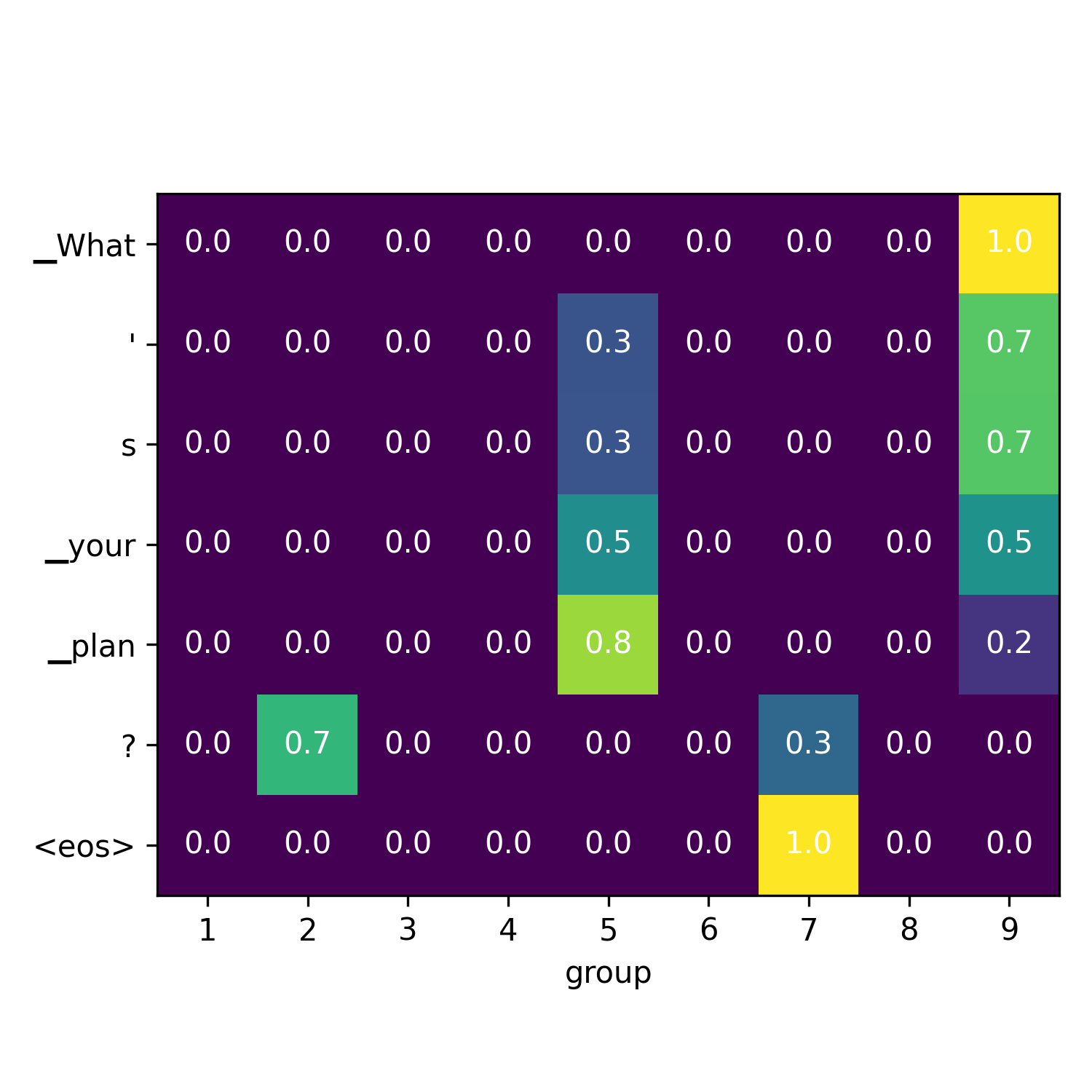}
        \caption{Latent Grouping}
        \label{fig:visualization-en-de-4-grouping}
    \end{subfigure}\hfill
    \begin{subfigure}{0.415\linewidth}
        \includegraphics[width=1\linewidth]{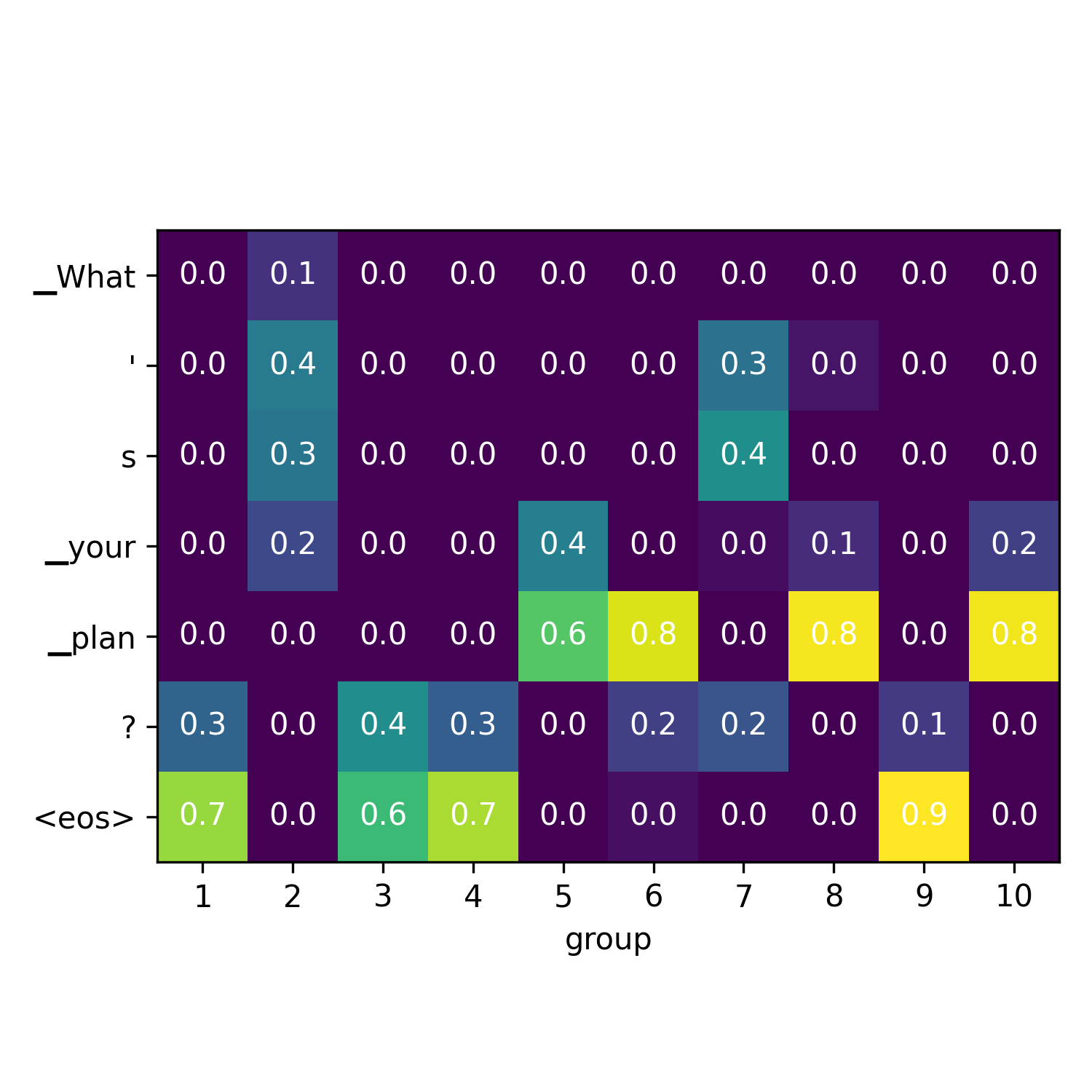}
        \caption{Latent Selecting}
        \label{fig:visualization-en-de-4-selecting}
    \end{subfigure}\hfill
    \begin{subfigure}{0.4\linewidth}
        \includegraphics[width=1\linewidth]{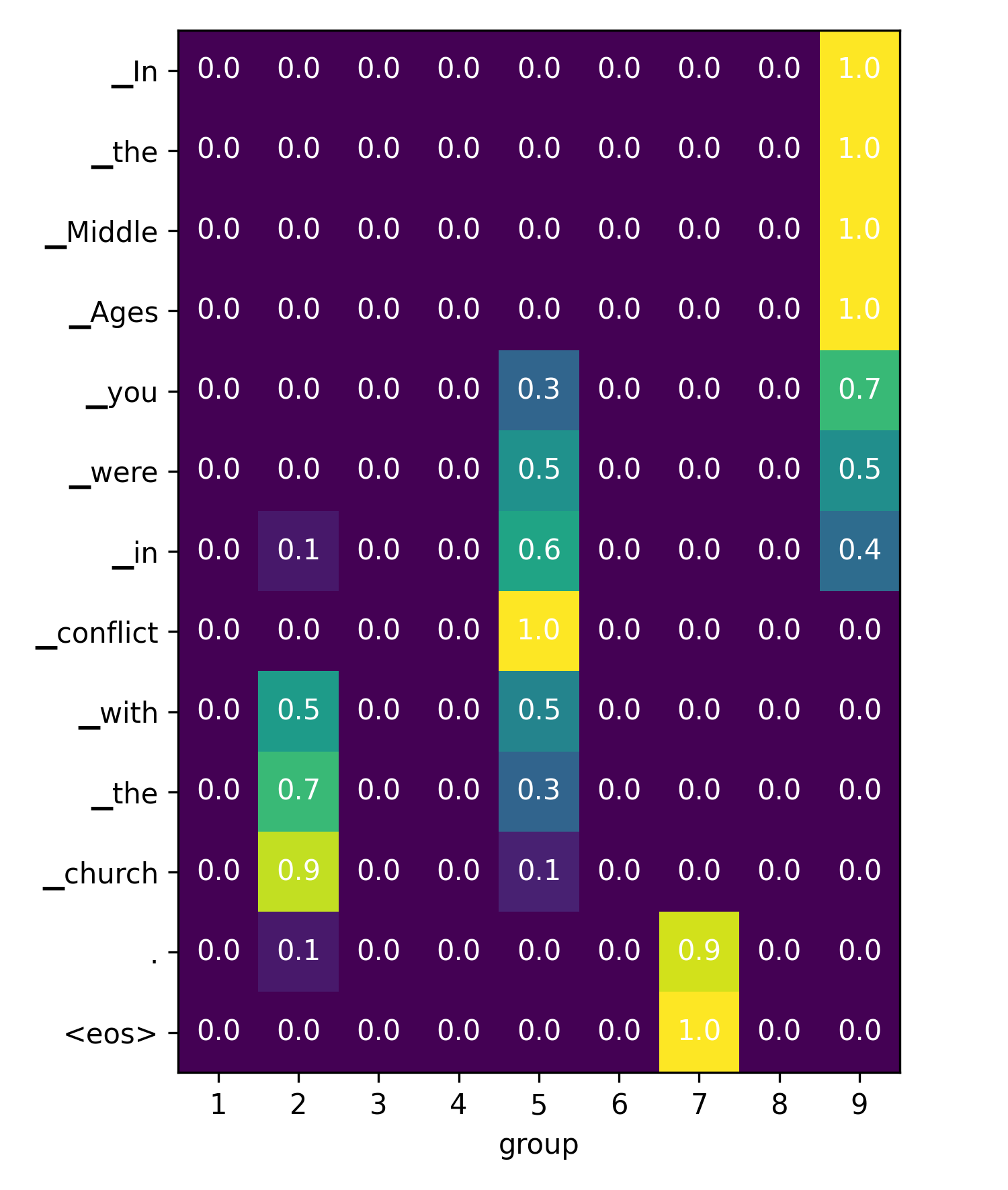}
        \caption{Latent Grouping}
        \label{fig:visualization-en-de-16-grouping}
    \end{subfigure}\hfill
    \begin{subfigure}{0.415\linewidth}
        \includegraphics[width=1\linewidth]{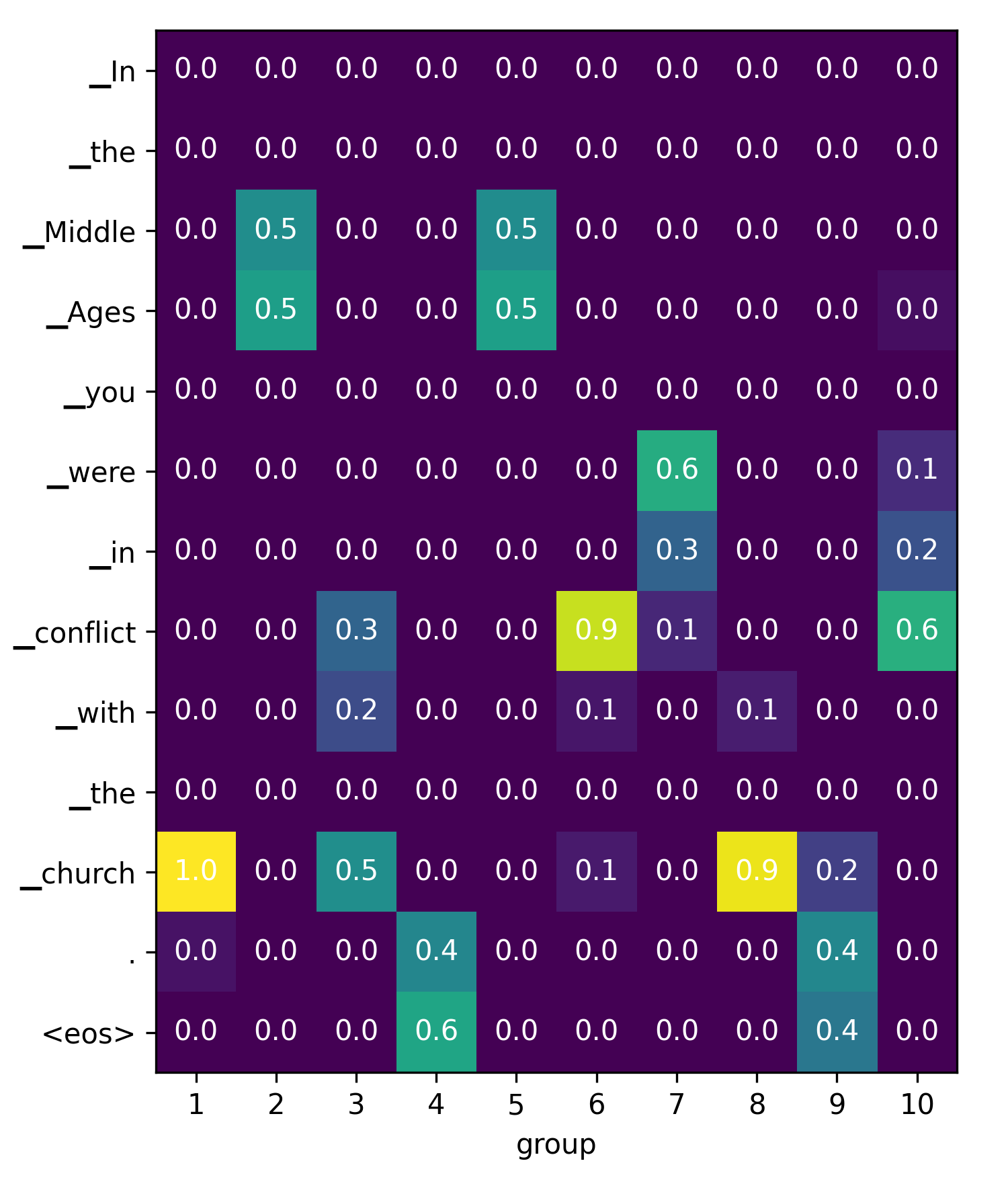}
        \caption{Latent Selecting}
        \label{fig:visualization-en-de-16-selecting}
    \end{subfigure}\hfill
    \begin{subfigure}{0.4\linewidth}
        \includegraphics[width=1\linewidth]{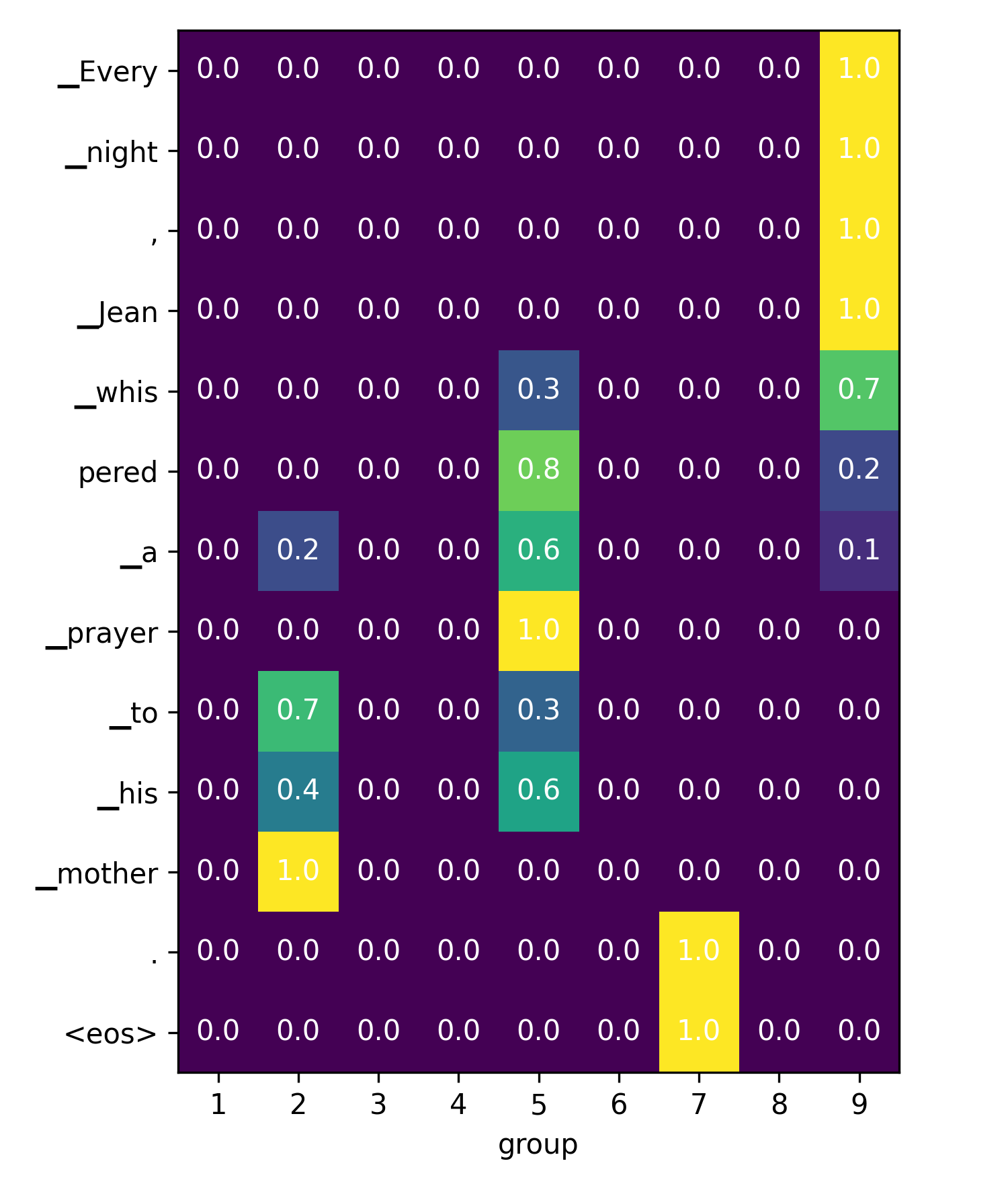}
        \caption{Latent Grouping}
        \label{fig:visualization-en-de-18-grouping}
    \end{subfigure}\hfill
    \begin{subfigure}{0.415\linewidth}
        \includegraphics[width=1\linewidth]{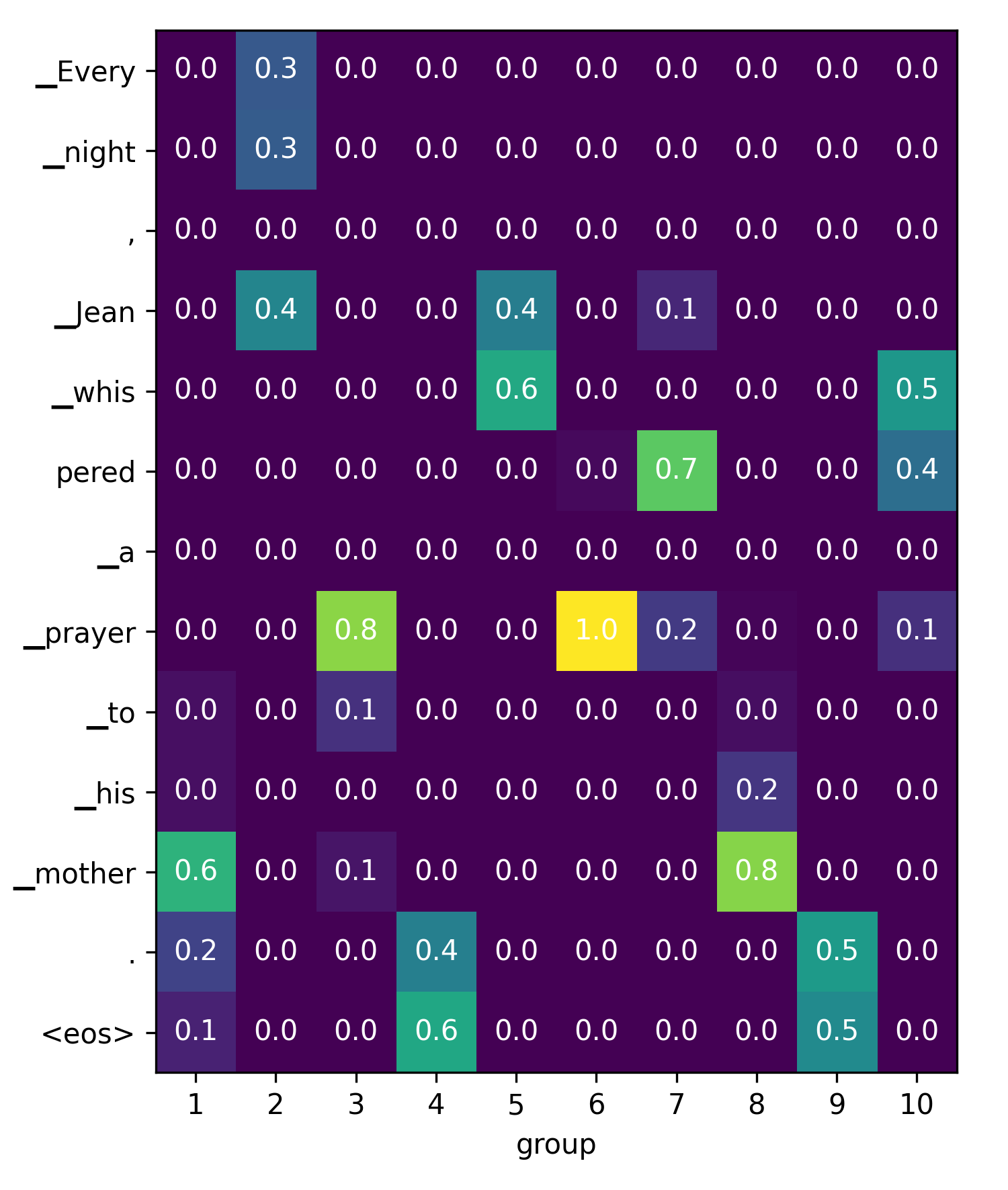}
        \caption{Latent Selecting}
        \label{fig:visualization-en-de-18-selecting}
    \end{subfigure}\hfill
    \caption{Visualization of tokens of the sentences from the ContraPro dataset \citep{muller-etal-2018-large} grouped (\ref{fig:visualization-en-de-4-grouping}, \ref{fig:visualization-en-de-16-grouping}, \ref{fig:visualization-en-de-18-grouping}) and selected (\ref{fig:visualization-en-de-4-selecting}, \ref{fig:visualization-en-de-16-selecting}, \ref{fig:visualization-en-de-18-selecting}) by the model using Latent Grouping and Latent Selecting.}
    \label{fig:visualization-en-de-extended}
\end{figure*}

\begin{figure*}[!ht]
\center{}
    \begin{subfigure}{0.4\linewidth}
        \includegraphics[width=1\linewidth]{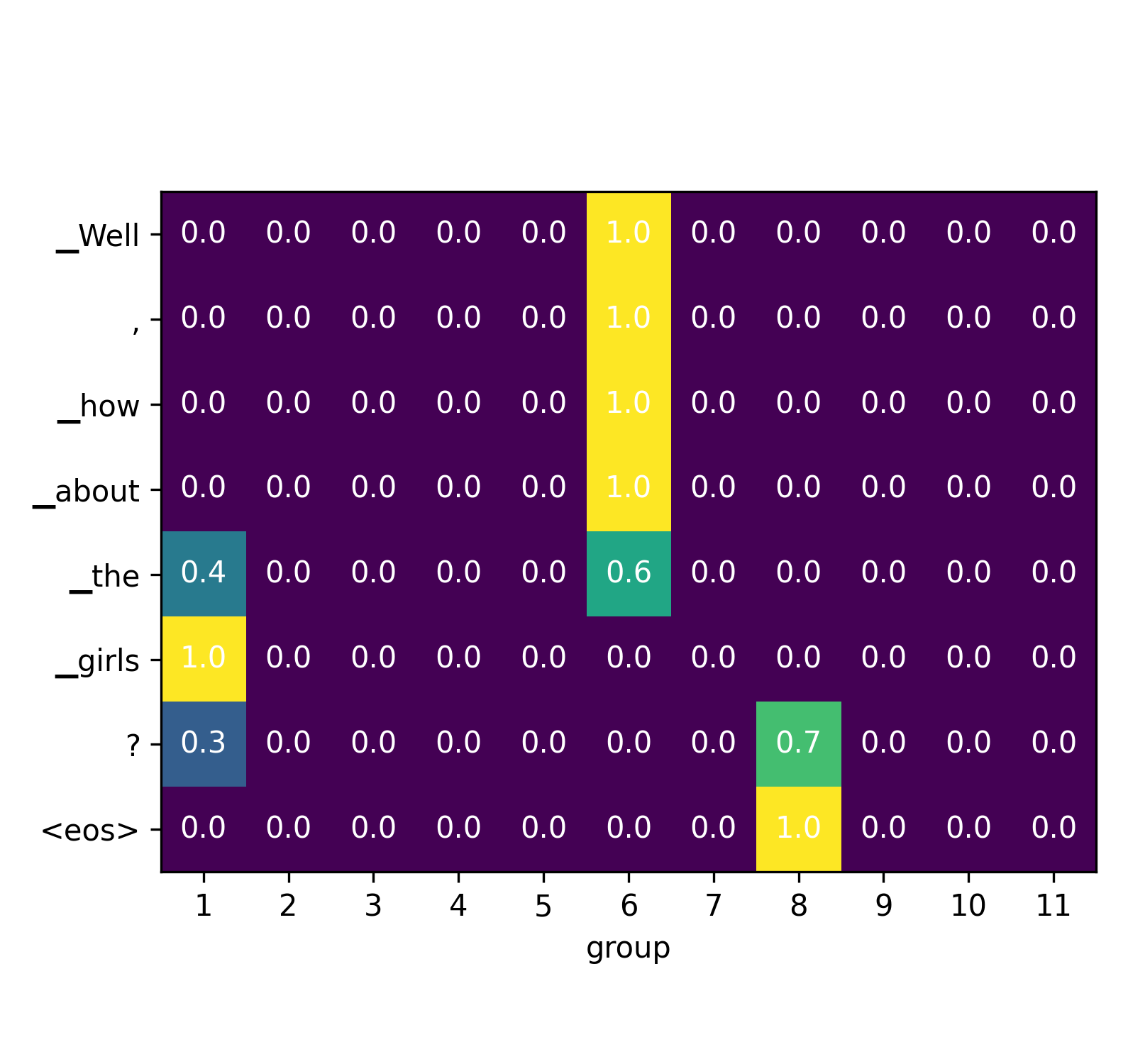}
        \caption{Latent Grouping}
        \label{fig:visualization-en-fr-18-grouping}
    \end{subfigure}\hfill
    \begin{subfigure}{0.415\linewidth}
        \includegraphics[width=1\linewidth]{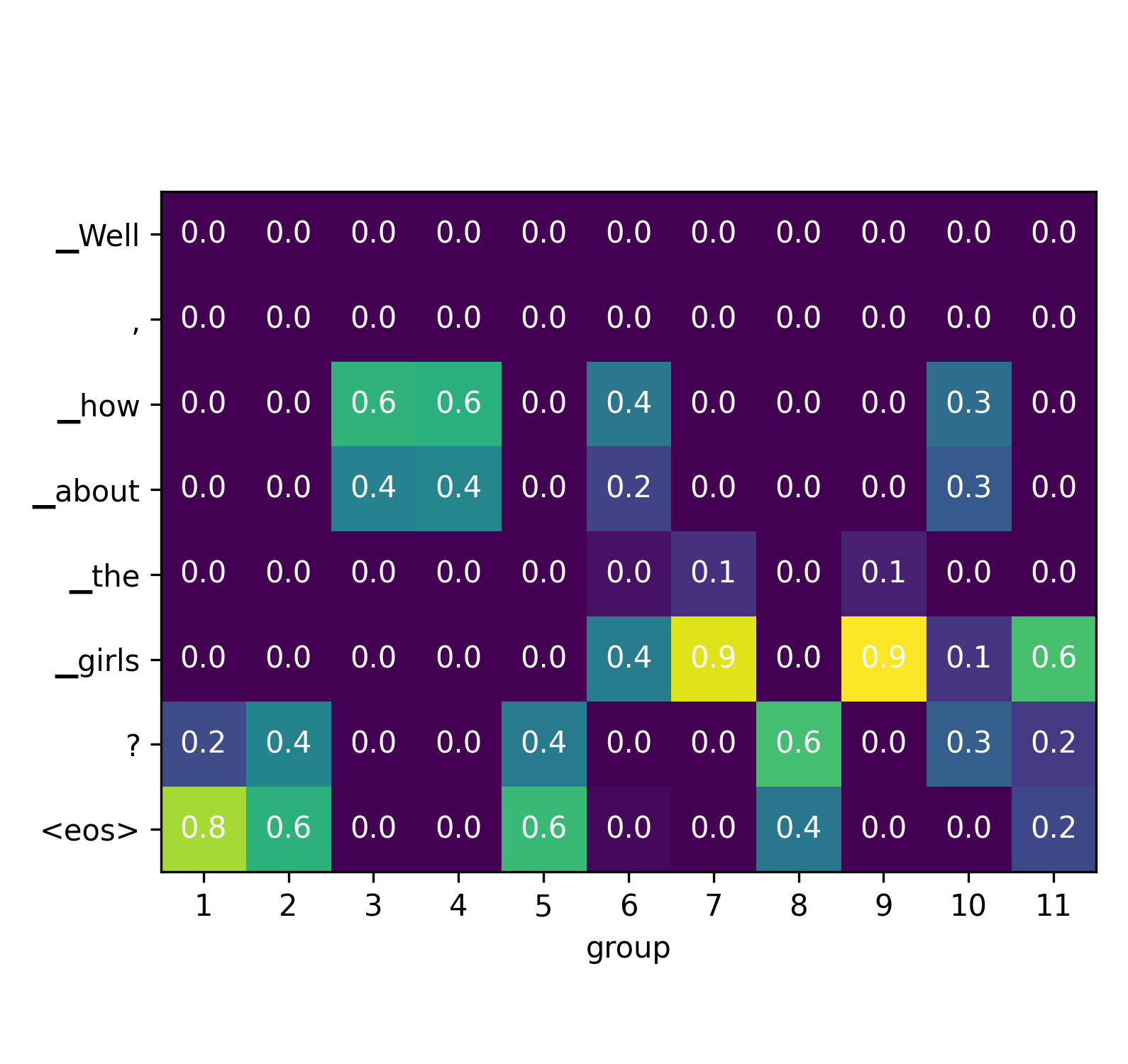}
        \caption{Latent Selecting}
        \label{fig:visualization-en-fr-18-selecting}
    \end{subfigure}\hfill
    \begin{subfigure}{0.4\linewidth}
        \includegraphics[width=1\linewidth]{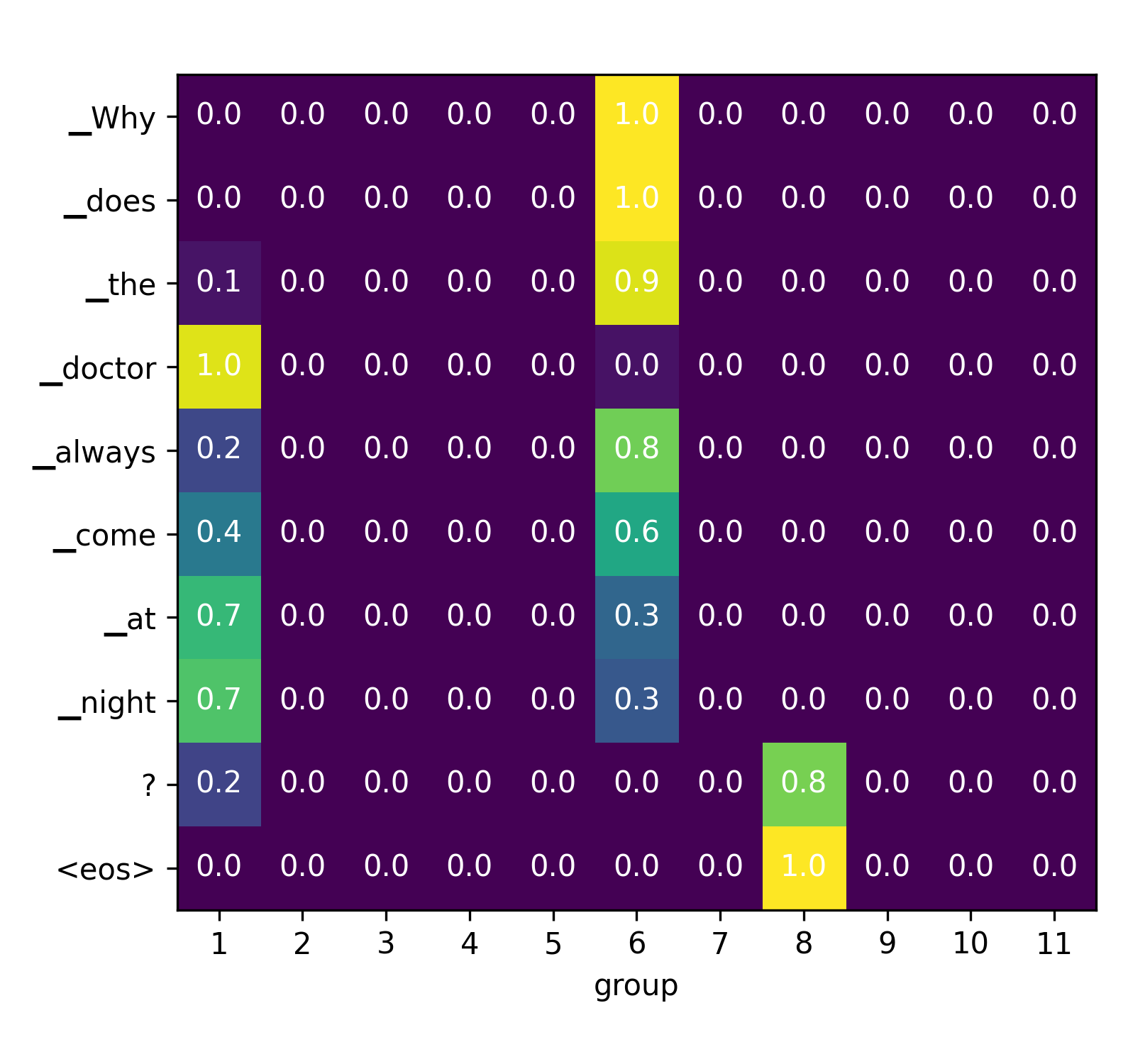}
        \caption{Latent Grouping}
        \label{fig:visualization-en-fr-21-grouping}
    \end{subfigure}\hfill
    \begin{subfigure}{0.415\linewidth}
        \includegraphics[width=1\linewidth]{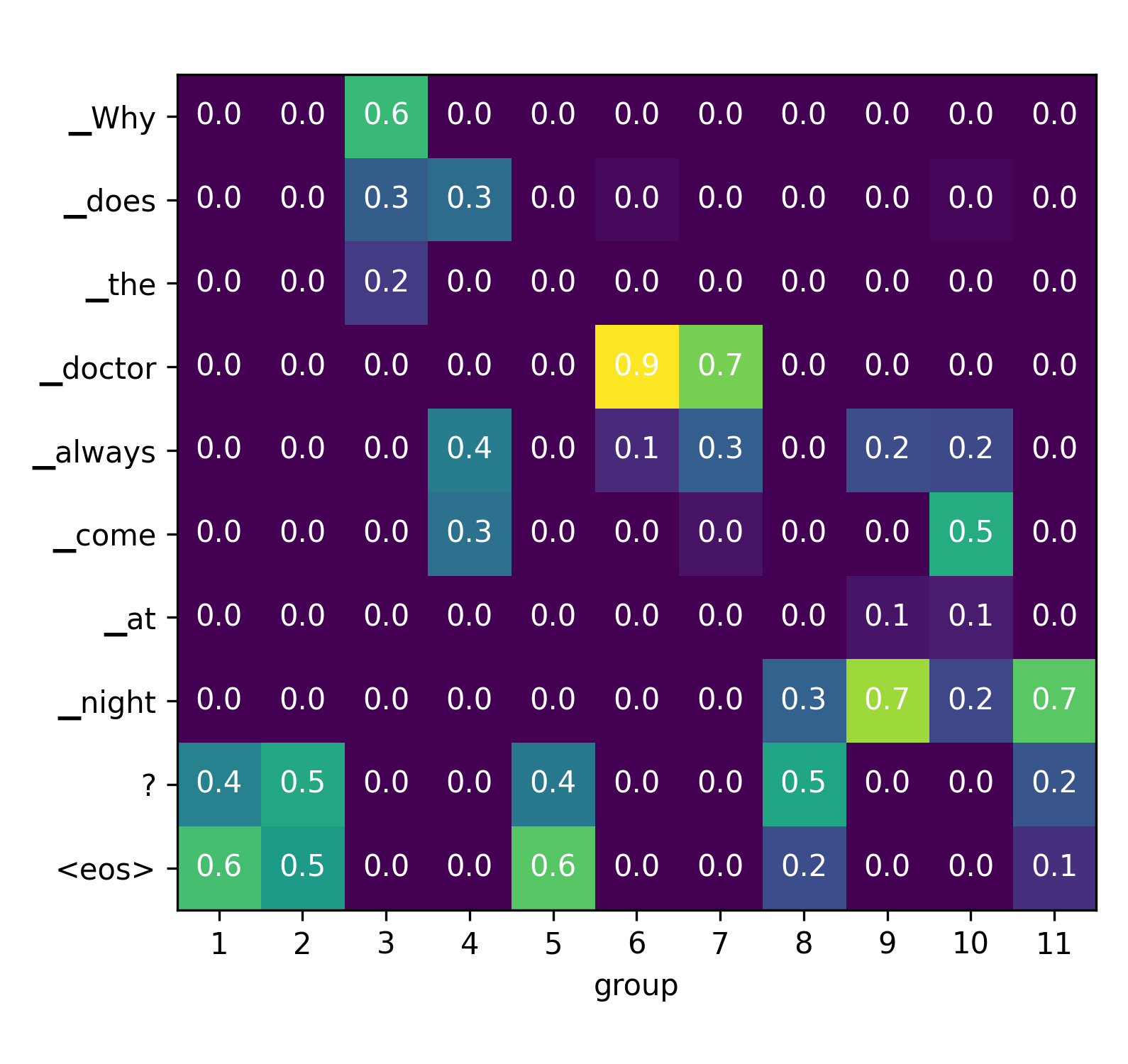}
        \caption{Latent Selecting}
        \label{fig:visualization-en-fr-21-selecting}
    \end{subfigure}\hfill
    \begin{subfigure}{0.4\linewidth}
        \includegraphics[width=1\linewidth]{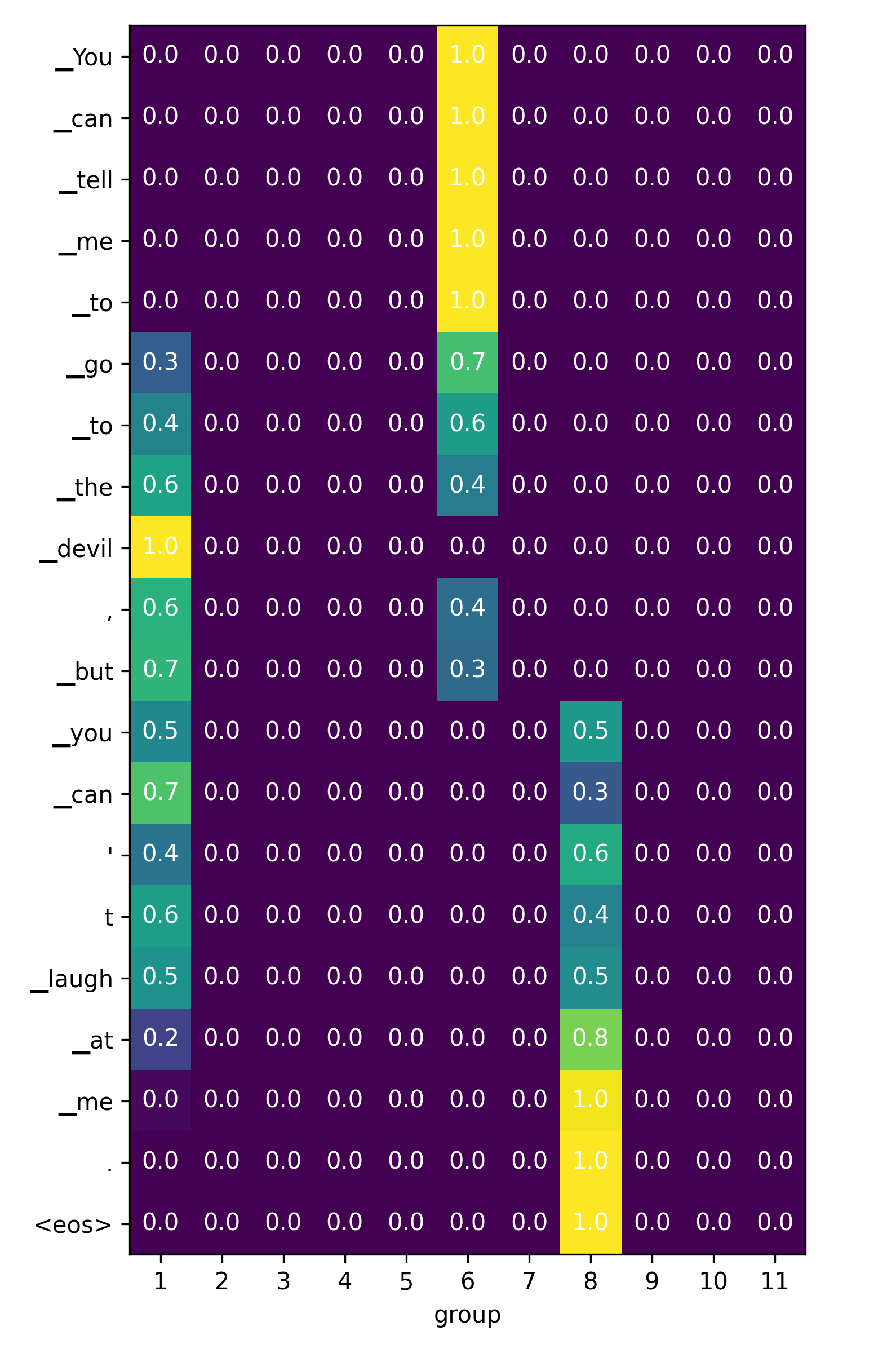}
        \caption{Latent Grouping}
        \label{fig:visualization-en-fr-19-grouping}
    \end{subfigure}\hfill
    \begin{subfigure}{0.415\linewidth}
        \includegraphics[width=1\linewidth]{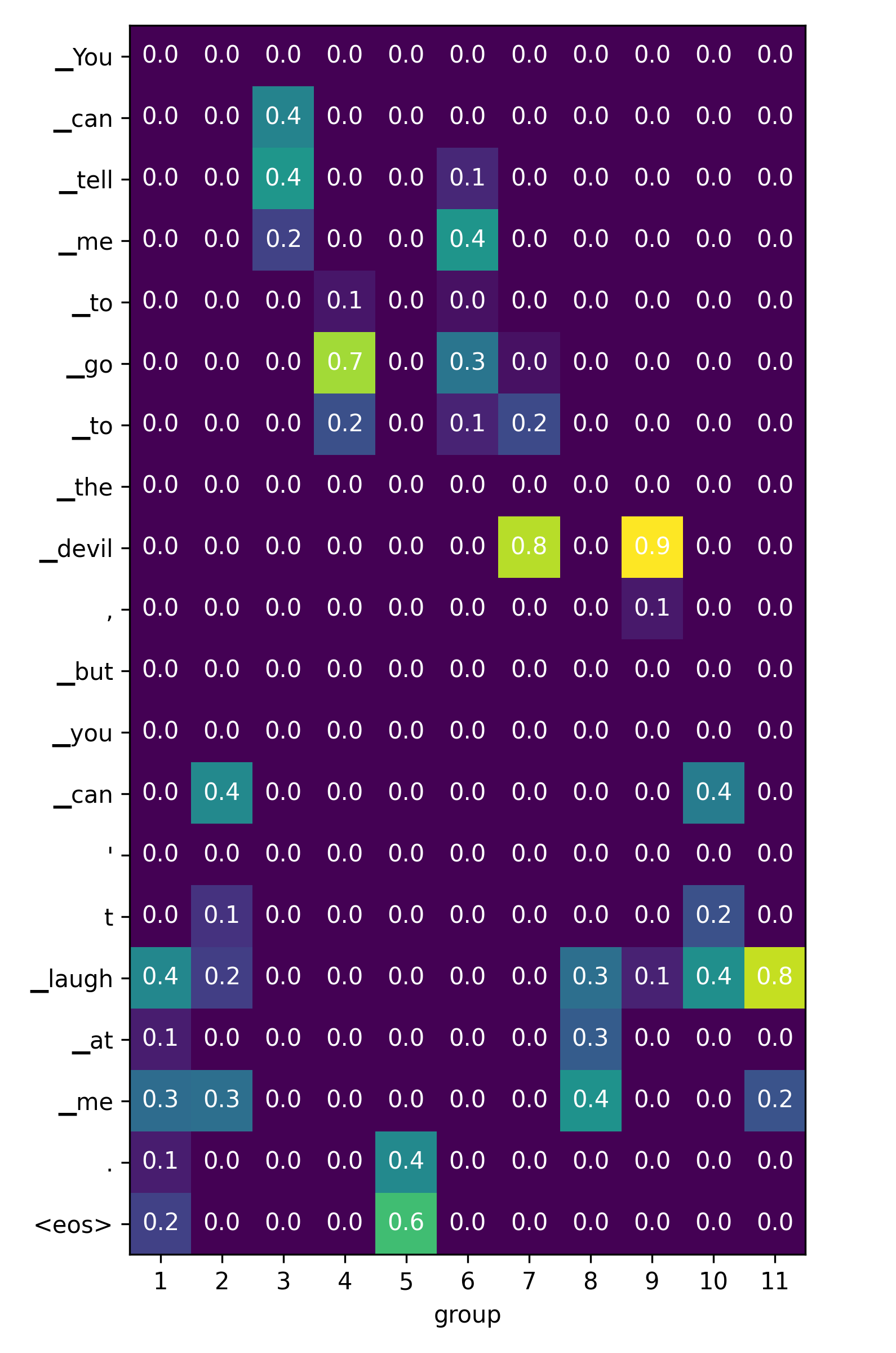}
        \caption{Latent Selecting}
        \label{fig:visualization-en-fr-19-selecting}
    \end{subfigure}\hfill
    \caption{Visualization of tokens of the sentences from the contrastive dataset by \citet{lopes-etal-2020-document} grouped (\ref{fig:visualization-en-fr-18-grouping}, \ref{fig:visualization-en-fr-21-grouping}, \ref{fig:visualization-en-fr-19-grouping}) and selected (\ref{fig:visualization-en-fr-18-selecting}, \ref{fig:visualization-en-fr-21-selecting}, \ref{fig:visualization-en-fr-19-selecting}) by the model using Latent Grouping and Latent Selecting.}
    \label{fig:visualization-en-fr-extended}
\end{figure*}

The visualizations of groupings and selections done by the models using Latent Grouping and Selecting of the additional examples from the ContraPro dataset \citep{muller-etal-2018-large} can be found in Figure~\ref{fig:visualization-en-de-extended}. Figure~\ref{fig:visualization-en-fr-extended} shows the visualizations of the groupings and selections of the sentences from the contrastive dataset by \citet{lopes-etal-2020-document}.

\end{document}